\documentclass[10pt,twocolumn,letterpaper]{article}

\usepackage{cvpr}              

\usepackage{times}
\usepackage{epsfig}
\usepackage{graphicx}
\usepackage{amsmath}
\usepackage{amssymb}
\usepackage{graphics} 
\usepackage{mathptmx} 
\usepackage{xcolor}  
\usepackage{soul}   
\usepackage{multirow}
\usepackage{multicol}
\usepackage{mathtools}
\usepackage{caption}
\usepackage{array}
\usepackage{bbding}
\usepackage[noend]{algpseudocode}
\usepackage{algorithm}
\usepackage{gensymb}
\usepackage{float}
\usepackage{makecell} 

\newcommand{\funcref}[1]{\hyperref[alg:ooam]{\texttt{#1}}}

\definecolor{cvprblue}{rgb}{0.21,0.49,0.74}
\usepackage[pagebackref,breaklinks,colorlinks,allcolors=cvprblue]{hyperref}

\def\paperID{44} 
\def\confName{CVPR}
\def\confYear{2025}

\title{SynTable: A Synthetic Data Generation Pipeline for Unseen Object Amodal Instance Segmentation of Cluttered Tabletop Scenes}

\author{Zhili Ng$^*$ \and Haozhe Wang$^{*,\dag}$ \and Zhengshen Zhang$^*$ \and Francis Eng Hock Tay \and Marcelo H. Ang Jr.\\
Advanced Robotics Centre, National University of Singapore\\
{\tt\small \{ng.zhili, wang\_haozhe, zhengshen\_zhang\}@u.nus.edu,} {\tt\small \{mpetayeh, mpeangh\}@nus.edu.sg}
}

\begin{document}
\maketitle
\def\thefootnote{*}\footnotetext{These authors contributed equally to this work.}
\def\thefootnote{\dag}\footnotetext{Corresponding author.}
\begin{abstract}
In this work, we present SynTable, a unified and flexible Python-based dataset generator built using NVIDIA's Isaac Sim Replicator Composer for generating high-quality synthetic datasets for unseen object amodal instance segmentation of cluttered tabletop scenes. Our dataset generation tool can render complex 3D scenes containing object meshes, materials, textures, lighting, and backgrounds. Metadata, such as modal and amodal instance segmentation masks, object amodal RGBA instances, occlusion masks, depth maps, bounding boxes, and material properties can be automatically generated to annotate the scene according to the users' requirements. Our tool eliminates the need for manual labeling in the dataset generation process while ensuring the quality and accuracy of the dataset. In this work, we discuss our design goals, framework architecture, and the performance of our tool. We demonstrate the use of a sample dataset generated using SynTable for training a state-of-the-art model, UOAIS-Net. Our state-of-the-art results show significantly improved performance in Sim-to-Real transfer when evaluated on the OSD-Amodal dataset. We offer this tool as an open-source, easy-to-use, photorealistic dataset generator for advancing research in deep learning and synthetic data generation. The links to our source code, demonstration video, and sample dataset can be found in the supplementary materials.


\end{abstract}
\vspace{-5mm}
\section{Introduction}
\label{sec:intro}

\begin{figure}[t]
	\centering
	\includegraphics[width=\linewidth]{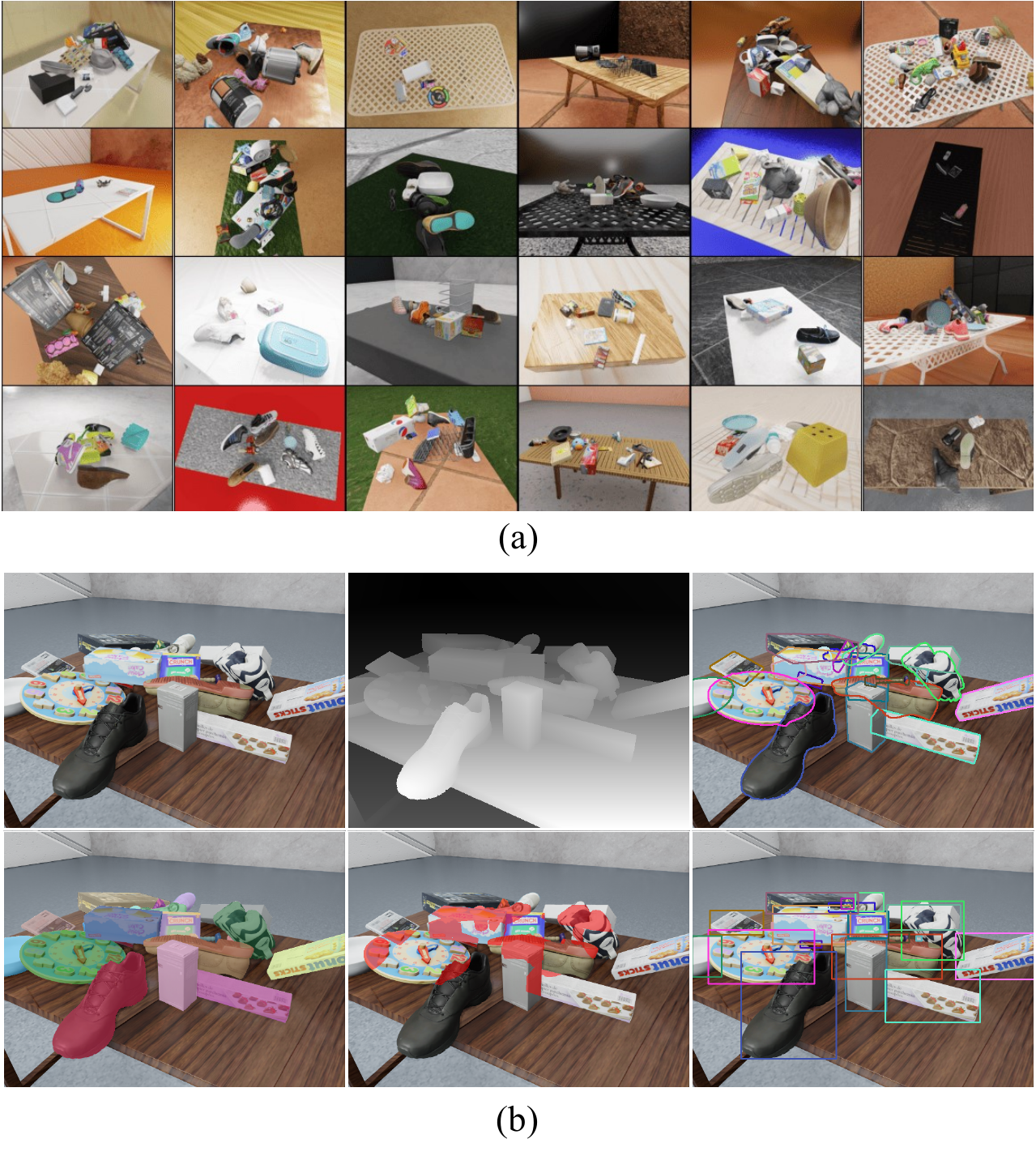}
	\caption{(a) RGB outputs of photorealistic cluttered tabletop scenes generated by SynTable pipeline. (b) Visualization of RGB Images, Depth Images, Object Amodal Masks, Object Visible Masks, Object Occlusion Masks, and Object Visible Bounding Boxes.}
	\label{teaser}
\vspace{-6mm}
\end{figure}


Amodal completion is a perceptual ability that enables the perception of whole objects, even when they are partially occluded \cite{Amodal_Instance_Segmentation, A_Survey}.  It encompasses three key tasks: amodal shape completion, amodal appearance completion and occlusion order. Amodal shape completion involves predicting the complete structure of an object beyond its visible portion, typically represented as a binary segmentation mask that includes both visible and occluded regions. Amodal appearance completion refers to the process of inferring the likely apperance of the hidden regions of an object based on its visible parts (RGB values of hidden pixels). Occlusion Order considers the occlusion relationship between objects, distinguishing between occluders (objects that obscure others) and occludees (objects being occluded), which can involve no occlusion or bi-directional occlusion. Humans are capable of ``filling in" the occluded appearance of invisible objects, owing to their vast experience in perceiving countless objects in various contexts and scenes. This ability to infer an object’s complete structure from its partial appearance is critical for systems requiring holistic scene understanding, such as augmented or virtual reality, and robotics and automation. In modern vision systems, accurately comprehending occluded objects in cluttered environments is essential for tasks ranging from object interaction to environment reconstruction.


There are three key challenges in amodal instance segmentation:
Firstly, the lack of large-scale, high-quality datasets for unseen object amodal instance segmentation (UOAIS) limits the performance of vision systems in real-world applications \cite{mti2030057}. While datasets exist for object detection and segmentation \cite{Jacquard,Efficient_grasping_from_RGBD_images,Dexnet2.0,8460609,GraspNet-1Billion}, only a few address UOAIS \cite{back2022unseen}. This is largely due to the difficulty of manually annotating amodal data, as human annotators must estimate occluded regions, leading to inherent subjectivity and inconsistencies in ground-truth annotations \cite{back2022unseen,KINS_Dataset,Semantic_Amodal_Segmentation}.

Secondly, synthetic datasets often suffer from visual domain mismatch due to non-photorealistic rendering or insufficient domain randomization \cite{TOD}, resulting in poor Sim-to-Real transfer. Existing tools prioritize rendering speed over photorealism, limiting their utility for training robust vision models, which results in a poor Sim-to-Real transfer that will inevitably reduce the performance of algorithms in real-world applications.

Thirdly, the lack of automated tools for generating amodal annotations and evaluating occlusion relationships hinders progress in this domain. Existing evaluation metrics focus on visible object regions but do not assess a model’s ability to infer occlusion order — a critical capability for systems operating in cluttered scenes. For example, understanding occlusion hierarchies enables sequential task planning and reduces errors caused by overlapping objects. However, manual annotation of such relationships is prohibitively time-consuming, necessitating simulation tools as a more cost-effective and accurate solution.

In this work, we address these challenges by developing SynTable, a unified Python-based tool for generating customizable, photorealistic datasets for UOAIS in cluttered scenes. While our experiments focus on tabletop environments (common in interaction tasks), our framework generalizes to diverse settings. SynTable integrates rendering and annotation into a single pipeline, allowing users to control scene complexity, object variety, and annotation types. Built on NVIDIA’s Isaac Sim Replicator Composer, it leverages high-fidelity ray tracing and domain randomization to bridge the Sim-to-Real gap.

Our key contributions are summarized as follows:

\begin{enumerate}
    \item We develop a pipeline to automatically render photorealistic cluttered tabletop scenes and generate ground truth amodal instance segmentation masks, eliminating manual labeling in dataset generation. Our designed dataset generation tool creates photorealistic and accurately-labeled custom datasets for UOAIS (refer to Figure \ref{teaser}(a)).
    \item Our tool provides a rich set of annotations related to amodal instance segmentation (refer to Figure \ref{teaser}(b)): modal (visible) and amodal instance segmentation masks, RGBA object instances, occlusion masks, occlusion rates, and occlusion order adjacency matrix. Users can easily select which annotations to include in their dataset based on the requirements of their application.
    \item We proposed a novel method to evaluate how accurately an amodal instance segmentation model can determine object occlusion ordering in a scene by computing the scene's Occlusion Order Accuracy ($ACC_{OO}$).
    \item We generated an open-sourced large-scale sample synthetic dataset using our tool consisting of amodal instance segmentation labels for users to train and evaluate amodal segmentation models on 1075 novel objects, designed to benchmark amodal segmentation in occlusion-rich scenarios.
\end{enumerate}

\section{Related Works}
\label{sec:related_works}

\subsection{Amodal Instance Segmentation in Vision Systems}
Recent advances in amodal instance segmentation aim to enhance object detection and tracking in complex scenes. However, challenges such as limited training data and Sim-to-Real gaps persist, particularly in cluttered environments where occlusion reasoning is critical.

\textbf{Lack of Large-scale High-quality Training Data.}
While datasets like \cite{DYCE,OLMD,CSD} have advanced amodal segmentation for indoor scenes, few address occlusion-rich scenarios in everyday interaction tasks. Existing efforts often focus on narrow domains: for example, \cite{MetaGraspNet} introduced a benchmark for multi-object interaction in industrial settings, but its limited scene and object diversity restrict broader applicability. Similarly, the Object Segmentation Database (OSD) \cite{OSD} and Object Cluttered Indoor Dataset (OCID) \cite{OCID} pioneered tools for segmentation in cluttered scenes but lack amodal annotations. Recent work by Back \textit{et al.} \cite{back2022unseen} manually added amodal masks to OSD, yet this approach remains labor-intensive and prone to human error.


\textbf{Sim-to-Real Problem.}
Synthetic datasets like the Tabletop Object Dataset (TOD) \cite{TOD} and UOAIS-Sim \cite{back2022unseen} struggle with photorealism and domain randomization, leading to significant Sim-to-Real gaps. For instance, TOD’s non-photorealistic rendering limits its utility for training models deployed in real-world applications such as augmented reality or autonomous navigation.


\subsection{Tools for Generating Synthetic Datasets}
With the rapid development of deep learning, the demand of researchers for synthetic datasets has increased in recent years, leading to the increased development of various tools for generating these datasets \cite{SAPIEN}. For robotics and computer vision applications, PyBullet and MuJoCo \cite{MuJoCo} are commonly used physical simulators to generate synthetic data. Xie \textit{et al.} \cite{TOD} pre-trained an RGB-D unseen object instance segmentation model using PyBullet. Tobin \textit{et al.} \cite{Domain_randomization} used MuJoCo to generate synthetic images with domain randomization, which can bridge the Sim-to-Real gap by realistically randomizing 3D content. Simulation tools such as PyBullet and MuJoCo typically come with renderers that are accessible and flexible, but they lack physically based light transport simulation, photorealism, material definitions, and camera effects. 

To obtain better rendering capabilities, researchers also explored the use of video game-based simulation tools, such as Unreal Engine (UE4) or Unity 3D. For example, Qiu and Yuille \cite{UnrealCV} exported specific metadata by adding a plugin to UE4. Besides, Unity 3D can generate metadata and produce scenes for computer vision applications using the official computer vision package. Although game engines provide the most advanced rendering technology, they prioritize frame rate over image quality and offer limited capabilities in light transport simulation. 

Ray-tracing technology has gained significant traction in creating photorealistic synthetic datasets, as it enables the simulation of light behavior with high accuracy. Software applications such as Blender, NVIDIA OptiX, and NVIDIA Isaac Sim have all incorporated ray-tracing techniques into their functionality. The Replicator Composer, a component of NVIDIA Isaac Sim, constitutes an excellent tool for creating tailored synthetic datasets to meet various requirements in robotics. In this work, we leverage this platform to design a customized pipeline to generate a synthetic dataset tailored to the specific demands of UOAIS for cluttered tabletop scenes.
\vspace{-2mm}
\section{Method}
\label{sec:method}


\begin{figure*}[t]
	\centering
	\includegraphics[width=\linewidth]{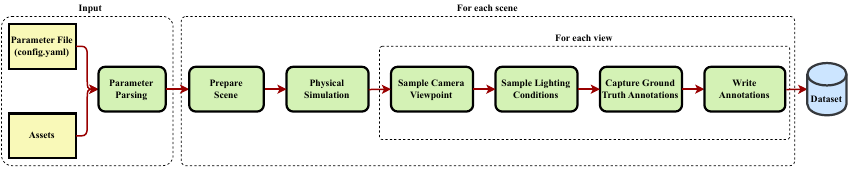}
	\caption{High-level overview of synthetic data generation pipeline.}
	\label{pipeline}	
\vspace{-4mm}
\end{figure*}
Our dataset generation pipeline is illustrated in Figure \ref{pipeline}. Parameters and configurations of the scenes to be rendered are defined in a parameter file. Objects, materials, and light sources used in our pipeline are referred to as assets. The scene is prepared by rendering a tabletop scene with floating objects in Isaac Sim. A physical simulation is run to drop the rendered objects onto the table. For every view within a scene, camera viewpoints and lighting conditions are re-sampled. Subsequently, the annotations are captured to create the dataset. We provide additional details about each step of our data generation pipeline in Section \ref{sec:supp_method} of our supplementary materials.


\subsection{Preparing Each Scene}
To prepare each scene, a table is randomly sampled and rendered in the center of a room, as shown in Figure \ref{prepare_scene}. The texture and materials of the table, ceiling, wall, and floor are randomized for domain randomization while objects are added with randomized coordinates and orientations. We randomly sample (with replacement) $N_{lower}$ to $N_{upper}$ number of objects for each scene. Objects are initialized with real-life dimensions, mass, collision properties, randomized rotations and coordinates, ensuring diverse object arrangements across scenes. Additional details about our scene preparation method can be found in Section \ref{sec:scene_preparation}.


\subsection{Physical Simulation of Each Scene}

Rendered objects are dropped onto the table through a physics simulation to ensure the random placement of objects in the scene. Objects that rebound off the tabletop surface and land beyond the spatial coordinate region of the tabletop surface are removed, excluding extraneous objects from annotations. We provide more details about our physical simulation in Section \ref{sec:physical_simulation}.

\subsection{Sampling of Camera Viewpoints}

To capture annotations for each scene from multiple viewpoints, we enhance the approach of Gilles \textit{et al.} \cite{metagraspnet2022} (which only uses fixed viewpoints) by capturing the $V$ number of viewpoints at random positions within custom radii of two concentric hemispheres of custom radii. The calculation of the Cartesian coordinates of each viewpoint can be found in \ref{sec:sample_camera_viewpoint} of our supplementary materials. Each viewpoint is oriented such that the camera looks directly at the center of the tabletop surface.

\subsection{Sampling of Lighting Conditions}

To simulate various indoor lighting conditions for each viewpoint, we resample $L$ spherical light sources using a method similar to Section \ref{sec:sample_camera_viewpoint}. Please refer to Section \ref{supp_sec:sample_lighting} in our supplementary materials for more details. In contrast to Back \textit{et al.}'s \cite{back2022unseen} approach of using point light sources, we use spherical light sources emitting light in all directions to mimic light bulbs. Furthermore, we uniformly sample the temperatures and intensity of the light sources. Users can customize the number of spherical light sources, as well as their intensities and temperatures.

\begin{figure*}[t]
	\centering
	\includegraphics[width=\linewidth]{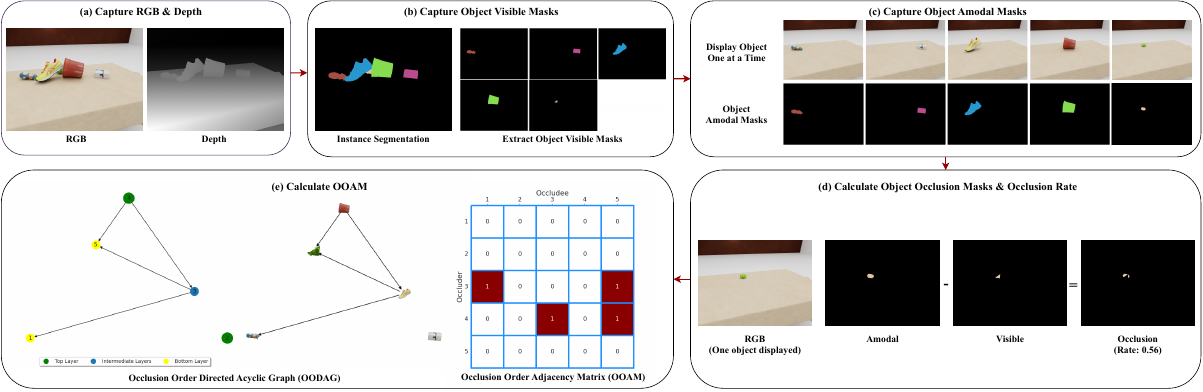} 
        \caption{The process of capturing annotations for a scene. For each viewpoint, (a) RGB and depth with all objects (b) object visible masks \& bounding box, (c) object amodal masks (including object amodal RGBA instances), (d) object occlusion masks and occlusion rate, (e) occlusion order adjacency matrix are captured.}
	\label{capture_annotations}
\end{figure*}
\subsection{Capturing of Ground Truth Annotations}
The process of capturing the annotations for a scene is illustrated in Figure \ref{capture_annotations}. In each view, the RGB and depth images of the tabletop scene will be captured (Figure \ref{capture_annotations}(a)). The built-in segmentation function in Isaac Sim Replicator Composer is used to capture the scene's instance segmentation mask from a viewpoint (Figure \ref{capture_annotations}(b)). Subsequently, the visible mask of each object is cropped from the scene's segmentation mask. 

For object amodal mask generation, we have developed the following steps.
Initially, all objects' visibility are disabled. For each object $o$ in the scene, its visibility is enabled and the instance segmentation function is utilized to capture its amodal mask and the amodal RGBA instance (Figure \ref{capture_annotations}(c)). We compute the object's occlusion mask and occlusion rate, as presented in (Figure \ref{capture_annotations}(d)). After capturing all object masks, we use Algorithm \ref{alg:ooam} to generate the Occlusion Order Adjacency Matrix (OOAM) for this viewpoint (Figure \ref{capture_annotations}(e)). For a scene with $M$ objects, the OOAM contains $M \times M$ elements, where the element ($i$, $j$) is a binary value in the matrix that indicates whether the object $i$ occludes the object $j$. Given the OOAM, we can easily construct the Occlusion Order Directed Graph (OODG) to visualize the occlusion order in the viewpoint (Figure \ref{capture_annotations}(e)). We provide a detailed explanation of the OODG in Section \ref{sec5} of our supplementary materials. After that, the visibility of all objects is enabled to prepare for the capturing of annotations from the next viewpoint of the scene.

\begin{algorithm}
\footnotesize
\captionsetup{font=footnotesize} 
\caption{A function to generate the OOAM of objects in a viewpoint.}\label{alg:ooam}
    \begin{algorithmic}[1]
    \Require Arrays of $visibleMasks$ and $occlusionMasks$ of objects in a scene
    \Ensure The OOAM of objects in a viewpoint
        \Function{generate\_ooam}{$visibleMasks$, $occlusionMasks$}
            \State Initialize OOAM as matrix of zeros
            \For{each \textbf{object} $\textbf{i}$ in length($VisibleMasks$) }
                \For{each \textbf{object} $\textbf{j}$ in length($OcclusionMasks$) }
                    \If{(\textbf{i} != \textbf{j}):}
                        \State $intersect =$ sum($visibleMasks$[\textbf{i}] $\cap$ $occlusionMasks$[\textbf{j}])
                        \If{$(intersect > 0):$}
                            \State OOAM[\textbf{i}][\textbf{j}] = 1
                        \EndIf
                    \EndIf
                \EndFor
            \EndFor
            \State return OOAM
        \EndFunction
        \State $\textbf{Note:}$ object \textbf{i} occludes object \textbf{j} if OOAM[\textbf{i}][\textbf{j}] = 1
    \end{algorithmic}
\end{algorithm}

\vspace{-2mm}


\section{Dataset Details}
\label{sec:dataset_details}

To demonstrate the capabilities of SynTable, we generated a sample synthetic dataset of cluttered tabletop scenes, SynTable-Sim, using our pipeline, to train and evaluate UOAIS models. Note that users can also generate other custom datasets that meet the specific requirements of their application using the SynTable pipeline.

\begin{table*}[t]
        \footnotesize
	\captionsetup{justification=centering}
	\renewcommand\arraystretch{1.0}
    \setlength\tabcolsep{6.5pt}
	\begin{center}
		\caption{A comparison of publicly available unseen object instance segmentation datasets for cluttered tabletop scenes. \textbf{$\#$} indicates the number of items. \textbf{VI}: Visible Instances. \textbf{OI}: Occluded Instances. \textbf{Avg. OR \%}: Avg. Occlusion Rate \%, i.e., the fraction of occluded pixels to amodal pixels across all object instances in the dataset.
  \textbf{AM}: Availability of amodal masks. \textbf{OM}: Availability of occlusion masks. \textbf{Order}: Availability of occlusion order relation information between objects. 
  \textbf{R/S}: Real or Synthetic. - indicates that the data was not available in the literature. \textbf{$^*$} indicates that the values were not provided in the original literature, but we were able to compute the values.}
		\label{comparing_different_datasets}
		\begin{tabular}{p{2.2cm}<{\centering}|p{1.15cm}<{\centering}|p{1.15cm}<{\centering}|p{1.15cm}<{\centering}|p{1.4cm}
  <{\centering}|p{1.15cm}<{\centering}|p{1.8cm}<{\centering}|p{0.55cm}<{\centering}|p{0.55cm}<{\centering}|p{0.65cm}<{\centering}|p{0.55cm}<{\centering}}
                \hline
                \hline
                Dataset & \#Images & \#Objects & \#Scenes & \#VI & \#OI & Avg. OR (\%) & AM & OM & Order & R/S\\ 
                \hline
                OCID \cite{OCID} & 2,390 & 89 & 96 & 19,097$^*$ & - & - & \textcolor{red}{\XSolidBrush} & \textcolor{red}{\XSolidBrush} &  \textcolor{red}{\XSolidBrush} &  R \\
                \hline
                OSD \cite{OSD} & 111 & - & 111 & 474$^*$ & - & - & \textcolor{red}{\XSolidBrush} & \textcolor{red}{\XSolidBrush} &  \textcolor{red}{\XSolidBrush} & R \\
                \hline
                OSD-Amodal \cite{back2022unseen} & 111 & - & 111 & $474^*$ & 237$^*$ & \textbf{24.11}$^*$ & \textcolor{green}{\CheckmarkBold} & \textcolor{green}{\CheckmarkBold} & \textcolor{red}{\XSolidBrush} & R \\
                \hline
                UOAIS-Sim & \multirow{2}*{25,000} & \multirow{2}*{375} & \multirow{2}*{500} & \multirow{2}*{$356,885^*$} & \multirow{2}*{$127,129^*$} & \multirow{2}*{$11.16^*$} & \multirow{2}*{\textcolor{green}{\CheckmarkBold}} & \multirow{2}*{\textcolor{green}{\CheckmarkBold}} & \multirow{2}*{\textcolor{red}{\XSolidBrush}} & \multirow{2}*{S} \\
                (Tabletop) \cite{back2022unseen} & & & & & & & & & &\\  
                \hline
                \textbf{SynTable-Sim} & \multirow{2}*{\textbf{50,000}} & \multirow{2}*{\textbf{1075}} & \multirow{2}*{\textbf{1000}} & \multirow{2}*{\textbf{744,454}} & \multirow{2}*{\textbf{482,921}} & \multirow{2}*{\underline{17.56}} & \multirow{2}*{\textcolor{green}{\CheckmarkBold}} & \multirow{2}*{\textcolor{green}{\CheckmarkBold}} & \multirow{2}*{\textcolor{green}{\CheckmarkBold}} & \multirow{2}*{S} \\
                \textbf{(Ours)} & & & & & & & & & &\\ 
                \hline
                \hline
            \end{tabular}
        \end{center}
\vspace{-4mm}
\end{table*}

\subsection{Object Models Used in Generating SynTable-Sim}
We use 1075 object CAD models from the Google Scanned Objects dataset \cite{downs2022google} and the Benchmark for the 6D Object Pose Estimation (BOP) \cite{hodan2018bop} to generate our train dataset. The Google Scanned Objects dataset features more than 1030 photorealistic 3D scanned household objects with real-life dimensions, and BOP features 3D object models from household and industrial objects. Upon inspection of the Google Scanned Objects dataset, we filter out invalid objects that contain more than two instances in each model and keep the remaining 891 valid objects for our training dataset. From the BOP, we exclude 21 objects from the YCB-Video dataset that we include in our validation dataset and use the remaining 184 objects for our training dataset. We also create a synthetic validation set using 78 novel objects from the YCB dataset \cite{7251504}. We sample a table object from 10 Omniverse Nucleus table assets to provide randomization for each scene. To load the 3D object models into Isaac Sim, we converted the OBJ and texture files to the Universal Scene Description (USD) format.

\subsection{Dataset Configuration}
With 50 viewpoints for each scene, we generated 900 scenes to create 45,000 RGB-D images for the training dataset and 100 scenes to create 5,000 RGB-D images for the validation dataset. $N_{lower}=1$ to $N_{upper}=40$ objects are rendered in randomly textured tabletop planes in each scene. We used 130 materials from Omniverse Nucleus material assets to be randomly applied on the walls, floor, and table for domain randomization purposes. $L_{lower}=0$ to $L_{upper}=2$ spherical lights are sampled for each scene. The viewpoint and lighting hemisphere parameters are automatically sampled based on the table dimensions. The camera parameters used are horizontal aperture: 2.63, vertical aperture: 1.96, and focal length: 1.88 to mimic the configuration of the RealSense LiDAR Camera L515. The rest of the parameters follow the default configurations of the pipeline.

\subsection{Syntable-Sim Versus Other Cluttered Tabletop Datasets}
We compare our SynTable-Sim dataset with several existing cluttered tabletop datasets in Table \ref{comparing_different_datasets}. Our tabletop dataset is the only one that provides complete annotations for all aspects of amodal instance segmentation. Furthermore, our dataset contains the most extensive variety of objects, the highest number of occlusion instances, and the second highest average occlusion rate — critical factors that significantly enhance the complexity and realism of training scenarios. These characteristics make our dataset very challenging for amodal instance segmentation tasks.

Additionally, SynTable-Sim exhibits a significantly higher proportion of heavily occluded objects in its training set compared to UOAIS-Sim, aligning more closely with the OSD-Amodal dataset, as shown in Figure \ref{histogram_occlusion_rate} in the supplementary materials. This high occlusion density ensures that models trained on our dataset generalize better to real-world cluttered environments. Moreover, the weakly connected component size, which quantifies the number of mutually overlapped regions per OODG and serves as a metric for scene complexity \cite{lee2022instaorder}, is consistently larger in SynTable-Sim compared to UOAIS-Sim (Figure \ref{histogram_number_of_regions_per_connected_component} in the supplementary materials). This indicates that our dataset presents significantly more intricate occlusion patterns, enabling amodal segmentation models to learn more robust occlusion reasoning capabilities. 

 \vspace{-2mm}
\section{Experiments}
\label{sec:experiments}

In this section, we present the results of our experiments aimed at evaluating the effectiveness of our dataset generation pipeline in producing synthetic datasets with good Sim-to-Real transfer performance. We used our SynTable-Sim sample dataset to train a state-of-the-art (SOTA) UOAIS model, UOAIS-Net \cite{back2022unseen}. UOAIS-Net is evaluated on the SynTable-Sim validation set and the OSD-Amodal \cite{back2022unseen} test set. To verify consistency of our results and further demonstrate the capability of SynTable to improve the performance of a variety of different UOAIS models, we also train and evaluate three other UOAIS models---Amodal MRCNN~\cite{AMRCNN_ORCNN}, ORCNN~\cite{AMRCNN_ORCNN}, ASN~\cite{ASN}---on the SynTable-Sim and OSD-Amodal datasets respectively.


\subsection{Training Strategy}
We train UOAIS-Net on the UOAIS-Sim tabletop and SynTable-Sim datasets using an NVIDIA Tesla V100 GPU with 16 GB of memory. For both datasets, we used 90\% of the images for training and 10\% for validation. To train UOAIS-Net using the UOAIS-Sim tabletop dataset, we use the same hyperparameters as Back \textit{et al.} \cite{back2022unseen}. To train UOAIS-Net with SynTable-Sim, we modified the depth range hyperparameter, which is used to preprocess input depth images. Specifically, we changed the range from the 2500 mm to 40000 mm range set by Back \textit{et al.} to a narrower range of 250 mm to 2500 mm. This adjustment is required because our dataset reflects real-world proportions and has a smaller depth range than the UOAIS-Sim dataset. We also use a similar training strategy to train Amodal MRCNN, ORCNN, and ASN.

\subsection{Evaluation Metrics}

We measure the performance of UOAIS-Net on the following traditional metrics \cite{TOD,Dave2019TowardsSA, segmentation_of_moving_objects}: Overlap P/R/F, Boundary P/R/F, and \textit{F@.75} for the amodal, visible, and invisible masks. Overlap P/R/F and Boundary P/R/F evaluate the whole area and the sharpness of the prediction, respectively, where P, R, and F are the precision, recall, and F-measure of instance masks after the Hungarian matching, respectively. \textit{F@.75} is the percentage of segmented objects with an Overlap F-measure greater than 0.75. We also report the accuracy ($\textit{ACC}_\mathcal{O}$) and F-measure ($\textit{F}_\mathcal{O}$) of occlusion classification, where $\textit{ACC}_\mathcal{O} = \frac{\delta}{\alpha}$, $\textit{F}_\mathcal{O} = \frac{2P_oR_o}{P_o+R_o}$, $P_o = \frac{\delta}{\beta}$, $R_o = \frac{\delta}{\gamma}$. $\alpha$ is the number of the matched instances after the Hungarian matching. $\beta$, $\gamma$, and $\delta$ are the number of occlusion predictions, ground truths, and correct predictions, respectively. We provide more details about the evaluation metrics in Section \ref{sec3} of our supplementary materials.

Due to the subjectivity of the invisible masks of objects, the evaluation of the performance of the UOAIS model solely based on the overlap and boundary \textit{P/R/F} of segmented objects may be inaccurate. The current UOAIS occlusion evaluation metrics measure how well the model can predict whether individual objects are occluded. However, these metrics neglect hierarchical occlusion relationships, which are crucial for systems requiring structured scene understanding. The Occlusion Order Adjacency Matrix (OOAM) encodes these relationships, and the derived Occlusion Order Directed Graph (OODG) enables applications such as sequencing interactions in cluttered environments (for example, retrieving obscured items) or rendering occluded objects in augmented reality. To quantify a model’s ability to infer occlusion hierarchies, we propose the Occlusion Order Accuracy Occlusion Order Accuracy ($ACC_{OO}$) metric as defined in Equation \ref{eqn:Occlusion Order Accuracy}.


\vspace{-4mm}
\begin{equation}
ACC_{OO}= \frac{ sum(similarityMatrix) - gtOOAMDiagonalSize}{gtOOAMSize - gtOOAMDiagonalSize}
\label{eqn:Occlusion Order Accuracy}
\end{equation}

In Equation \ref{eqn:Occlusion Order Accuracy}, $similarityMatrix$ is the element-wise equality comparison between the ground truth OOAM, $gtOOAM$, and the predicted OOAM, $predOOAM$. As an object cannot occlude itself, the diagonal of any OOAM is always 0. Thus, we subtract the number of elements along the diagonal of $gtOOAM$, $gtOOAMDiagonalSize$, from the calculation of $ACC_{OO}$. $ACC_{OO}$ is used to evaluate the model's ability to accurately determine the order of occlusions in a clutter of objects by comparing the OOAM generated by the model to the ground truth OOAM using Algorithm \ref{alg:occlusion_order_accuracy}. We give a specific example of how to compute $ACC_{OO}$ in Sections \ref{sec4} and \ref{sec5} of our supplementary materials.

\begin{algorithm}
    \footnotesize
    \captionsetup{font=footnotesize} 
    \caption{Evaluating Occlusion Ordering Accuracy}
    \label{alg:occlusion_order_accuracy}
    \begin{algorithmic}[1]
    \Require The arrays of the ground truth and predicted visible and occlusion masks ($gtVisible$, $gtOcclusion$, $predVisible$, $predOcclusion$) 
    \Ensure  Scene occlusion order accuracy $ACC_{oo}$
        \State $gtOOAM = \funcref{GENERATE\_OOAM}(gtVisible, gtOcclusion)$
        \State Get groundtruth-prediction assignment pairs after Hungarian matching
        \State Extract $predVisible$ and $predOcclusion$ masks from assignment pairs
        \State $predOOAM = \funcref{GENERATE\_OOAM}(predVisible, predOcclusion)$
        \State $similarityMatrix = (predOOAM == gtOOAM)$ \Comment{Compare the similarity between the predicted and ground truth OOAMs}
        \State Calculate $ACC_{oo}$ using Equation \ref{eqn:Occlusion Order Accuracy}
    \end{algorithmic}
\end{algorithm}
\vspace{-4mm}

\subsection{Results}
\begin{table*}[!ht]
        \scriptsize
	\captionsetup{justification=centering}
	\renewcommand\arraystretch{1.0}
        \setlength\tabcolsep{6pt}
	\begin{center}
	\caption{The performance of UOAIS-Net on the \textbf{OSD-Amodal dataset} after training on the UOAIS-Sim and SynTable-Sim datasets. UOAIS-Net is trained with RGB-D images. \textbf{CR}: Crop Ratio lower bound. \textbf{HF}: Horizontal Flip. \textbf{CA}: Colour Augmentation. \textbf{PD}: Perlin Distortion. \textbf{OV}: Overlap F-measure, \textbf{BO}: Boundary F-measure, \textbf{F@.75}: Percentage of segmented objects with an Overlap F-measure greater than 0.75, \textbf{$\textit{F}_\mathcal{O}$}: Occlusion F-Measure, \textbf{$\text{ACC}_{\text{OO}}$}: Occlusion Order Accuracy}	
		\label{eval_on_OSD_amodal}
		\begin{tabular}{p{0.2cm}<{\centering}|p{2.5cm}<{\centering}|p{0.2cm}<{\centering}|p{0.2cm}<{\centering}|p{0.2cm}<{\centering}|p{0.2cm}<{\centering}|p{0.4cm}<{\centering}|p{0.4cm}<{\centering}|p{0.6cm}<{\centering}|p{0.4cm}<{\centering}|p{0.4cm}<{\centering}|p{0.6cm}<{\centering}|p{0.4cm}<{\centering}|p{0.6cm}<{\centering}|p{0.4cm}<{\centering}|p{0.4cm}<{\centering}|p{0.6cm}<{\centering}|p{0.6cm}<{\centering}}
                \hline
                \hline
                \multirow{2}*{No.}&\multirow{2}*{Training Set}& \multicolumn{4}{c|}{Augmentation} & \multicolumn{3}{c|}{Amodal Mask} & \multicolumn{3}{c|}{Invisible Mask} & \multicolumn{2}{c|}{Occlusion} & \multicolumn{3}{c|}{Visible Mask} & \multirow{2}*{$ACC_{OO}$} \\
                &\multicolumn{1}{c|}{} &  \multicolumn{1}{c}{CR} & \multicolumn{1}{c}{HF} &\multicolumn{1}{c}{CA} & PD & \multicolumn{1}{c}{OV} & \multicolumn{1}{c}{BO} & \textit{F@.75} & \multicolumn{1}{c}{OV} & \multicolumn{1}{c}{BO} & \textit{F@.75} & \multicolumn{1}{c}{$\textit{F}_\mathcal{O}$} & $\textit{ACC}_\mathcal{O}$ & \multicolumn{1}{c}{OV} & \multicolumn{1}{c}{BO} & \textit{F@.75} &  \\
                \hline
                \multirow{1}*{1} & \multirow{1}*{UOAIS-Sim (Tabletop)} & \textcolor{red}{\XSolidBrush} & \textcolor{red}{\XSolidBrush} & \textcolor{red}{\XSolidBrush} & \textcolor{red}{\XSolidBrush} & 42.4 & 34.1 & 47.1 & 21.6 & 15.2 & 18.5 &  43.1 & 61.8 & 42.5 & 32.3 & 37.1 & 12.7 \\
                \hline
                \multirow{1}*{1} & \multirow{1}*{SynTable-Sim (Ours)} & \textcolor{red}{\XSolidBrush} & \textcolor{red}{\XSolidBrush} & \textcolor{red}{\XSolidBrush} & \textcolor{red}{\XSolidBrush} & \textbf{80.9} & 61.8 & \textbf{78.1} & \textbf{52.4} & \textbf{31.2} & 41.3 & \textbf{75.7} & \textbf{86.7} & \textbf{81.1} & 64.3 & \textbf{74.4} & \textbf{82.9} \\
                \hline
                \multirow{1}*{1} & \multirow{1}*{SynTable-Sim-0.5X (Ours)} & \textcolor{red}{\XSolidBrush} & \textcolor{red}{\XSolidBrush} & \textcolor{red}{\XSolidBrush} & \textcolor{red}{\XSolidBrush} & 80.7 & \textbf{63.8} & 77.3 & 51.9 & 30.2 & \textbf{42.9} & \textbf{75.7} & 84.1 & 80.5 & \textbf{65.4} & 71.7 & 82.7 \\
                \hline 
                \multicolumn{15}{c}{}\\[-0.5em]
                \hline
                \multirow{1}*{2} & \multirow{1}*{UOAIS-Sim (Tabletop)} & 0.8 & \textcolor{green}{\CheckmarkBold} & \textcolor{red}{\XSolidBrush} & \textcolor{red}{\XSolidBrush} & 26.1 & 33.1 & 66.7 & 15.5 & 7.7 & 20.4 & 60.8 & 78.1 & 25.9 & 27.6 & 51.8 & 42.7 \\
                \hline
                \multirow{1}*{2} & \multirow{1}*{SynTable-Sim (Ours)} & 0.8 & \textcolor{green}{\CheckmarkBold} & \textcolor{red}{\XSolidBrush} & \textcolor{red}{\XSolidBrush} & 67.7 & 56.0 & 81.2 & 49.4 & 30.1 & \textbf{48.6} & 72.5 & 89.8 & 71.8 & 61.3 & 78.2 & 86.6 \\
                \hline
                \multirow{1}*{2} & \multirow{1}*{SynTable-Sim-0.5X (Ours)} & 0.8 & \textcolor{green}{\CheckmarkBold} & \textcolor{red}{\XSolidBrush} & \textcolor{red}{\XSolidBrush} & \textbf{75.6} & \textbf{61.2} & \textbf{83.5} & \textbf{53.6} & \textbf{31.2} & 48.5 & \textbf{75.5} & \textbf{90.1} & \textbf{76.8} & \textbf{64.5} & \textbf{78.3} & \textbf{87.0} \\
                \hline 
                \multicolumn{15}{c}{}\\[-0.5em]
                \hline
                \multirow{1}*{3} & \multirow{1}*{UOAIS-Sim (Tabletop)} & 0.8 & \textcolor{green}{\CheckmarkBold} & \textcolor{green}{\CheckmarkBold} & \textcolor{green}{\CheckmarkBold} & 71.8 & \textbf{62.8} & 81.4 & \textbf{55.6} & \textbf{31.3} & \textbf{44.6} & \textbf{75.1} & 86.2  &  70.2 & \textbf{63.2} & 73.2 & 79.6 \\
                \hline
                \multirow{1}*{3} & \multirow{1}*{SynTable-Sim (Ours)} & 0.8 & \textcolor{green}{\CheckmarkBold} & \textcolor{green}{\CheckmarkBold} & \textcolor{green}{\CheckmarkBold} & \textbf{78.3} & 58.8 & 81.9 & 54.0 & 29.7 & 43.9 & 66.6 & 93.2 & \textbf{79.2} & 60.4 & 77.2 & \textbf{87.7} \\ 
                \hline
                \multirow{1}*{3} & \multirow{1}*{SynTable-Sim-0.5X (Ours)} & 0.8 & \textcolor{green}{\CheckmarkBold} & \textcolor{green}{\CheckmarkBold} & \textcolor{green}{\CheckmarkBold} & 74.0 & 57.5 & \textbf{83.3} & 49.2 & 23.9 & 41.0 & 65.7 & \textbf{93.4} & 74.2 & 59.2 & \textbf{79.2} & 87.6 \\
                \hline 
                \multicolumn{15}{c}{}\\[-0.5em]
                \hline
                \multirow{1}*{4} & \multirow{1}*{UOAIS-Sim (Tabletop)} & 0.5 & \textcolor{green}{\CheckmarkBold} & \textcolor{green}{\CheckmarkBold} & \textcolor{green}{\CheckmarkBold} & 49.0 & 50.3 & 82.7 & 42.3 & 23.9 & 40.3 & \textbf{68.9} & 84.0 & 47.3 & 50.0 & 70.6 & 80.4 \\
                \hline
                \multirow{1}*{4} & \multirow{1}*{SynTable-Sim (Ours)} & 0.5 & \textcolor{green}{\CheckmarkBold} & \textcolor{green}{\CheckmarkBold} &\textcolor{green}{\CheckmarkBold} & \textbf{64.4} & \textbf{51.5} & 84.3 & \textbf{47.3} & \textbf{24.2} & 47.4 & 60.0 & \textbf{91.9} & \textbf{65.3} & \textbf{53.7} & \textbf{78.2} & 87.0 \\
                 \hline
                \multirow{1}*{4} & \multirow{1}*{SynTable-Sim-0.5X (Ours)} & 0.5 & \textcolor{green}{\CheckmarkBold} & \textcolor{green}{\CheckmarkBold} &\textcolor{green}{\CheckmarkBold} & 55.0 & 47.2 & \textbf{85.9} & 43.2 & 22.0 & \textbf{48.4} & 55.4 & 91.5  &  55.3 & 46.6 & 76.9 & \textbf{87.8} \\
                \hline 
                \hline
            \end{tabular}
        \end{center}
\vspace{-4mm}
\end{table*}

\begin{table*}[!ht]
        \scriptsize
	\captionsetup{justification=centering}
	\renewcommand\arraystretch{1.0}
        \setlength\tabcolsep{0.8pt}
	\begin{center}
		\caption{A breakdown of the evaluation results of UOAIS-Net on the \textbf{OSD-Amodal dataset} for the first set of experiments after training on the UOAIS-Sim and SynTable-Sim dataset. \textbf{P}: Precision, \textbf{R}: Recall, \textbf{F}: F-measure, \textbf{F@.75}: Percentage of segmented objects with an Overlap F-measure greater than 0.75, \textbf{$\textit{F}_\mathcal{O}$}: Occlusion F-Measure, \textbf{$\text{ACC}_{\text{OO}}$}: Occlusion Order Accuracy}
		\label{comparing_PRF_on_OSD_amodal_1}
		\begin{tabular}{p{1.8cm}<{\centering}|
                            p{0.55cm}<{\centering}|p{0.55cm}<{\centering}|p{0.55cm}<{\centering}|
                            p{0.55cm}<{\centering}|p{0.55cm}<{\centering}|p{0.55cm}<{\centering}|p{0.75cm}<{\centering}|
                            p{0.55cm}<{\centering}|p{0.55cm}<{\centering}|p{0.55cm}<{\centering}|
                            p{0.55cm}<{\centering}|p{0.55cm}<{\centering}|p{0.55cm}<{\centering}|p{0.75cm}<{\centering}|
                            p{0.55cm}<{\centering}|p{0.55cm}<{\centering}|p{0.55cm}<{\centering}|
                            p{0.55cm}<{\centering}|p{0.55cm}<{\centering}|p{0.55cm}<{\centering}|p{0.75cm}<{\centering}|
                            p{0.55cm}<{\centering}|p{0.55cm}<{\centering}|p{0.55cm}<{\centering}|
                            p{0.55cm}<{\centering}|p{0.55cm}<{\centering}|p{0.55cm}<{\centering}}
                \hline
                \hline
                \multirow{3}*{Training Set} & \multicolumn{7}{c|}{Amodal Mask} & \multicolumn{7}{c|}{Invisible Mask}& \multicolumn{7}{c|}{Visible Mask} & \multicolumn{2}{c|}{Occlusion} & \multicolumn{1}{c}{\multirow{3}*{$ACC_{OO}$}}\\
                
                \cline{2-24}
                & \multicolumn{3}{c|}{Overlap} & \multicolumn{3}{c|}{Boundary} & \multirow{2}*{\textit{F@.75}} & \multicolumn{3}{c|}{Overlap} & \multicolumn{3}{c|}{Boundary} & \multirow{2}*{\textit{F@.75}} 
                 & \multicolumn{3}{c|}{Overlap} & \multicolumn{3}{c|}{Boundary} & \multirow{2}*{\textit{F@.75}} 
                 &\multicolumn{1}{c|}{\multirow{2}*{$\textit{F}_\mathcal{O}$}} &\multicolumn{1}{c|}{\multirow{2}*{$\textit{ACC}_\mathcal{O}$}} \\
                 
                & \multicolumn{1}{c}{P} & \multicolumn{1}{c}{R} & F & \multicolumn{1}{c}{P} & \multicolumn{1}{c}{R} & F & & \multicolumn{1}{c}{P} & \multicolumn{1}{c}{R} & F & \multicolumn{1}{c}{P} & \multicolumn{1}{c}{R} & F & & \multicolumn{1}{c}{P} & \multicolumn{1}{c}{R} & F & \multicolumn{1}{c}{P} & \multicolumn{1}{c}{R} & F & &
                  & \multicolumn{1}{c|}{}\\
                
                \hline
                 UOAIS-Sim & \multirow{2}*{35.9} &  \multirow{2}*{65.4} & \multirow{2}*{42.4} &  \multirow{2}*{31.4} &  \multirow{2}*{42.8} &  \multirow{2}*{34.1} &  \multirow{2}*{47.1} &  \multirow{2}*{55.9} &  \multirow{2}*{24.5} &  \multirow{2}*{21.6} &  \multirow{2}*{\textbf{45.3}} &  \multirow{2}*{19.3} &  \multirow{2}*{15.2} &  \multirow{2}*{18.5} &  \multirow{2}*{36.2} & \multirow{2}*{61.3} & \multirow{2}*{42.5} &  \multirow{2}*{30.8} & \multirow{2}*{39.2} & \multirow{2}*{32.3} & \multirow{2}*{37.1} & \multirow{2}*{43.1} & \multirow{2}*{61.8} & \multicolumn{1}{c}{\multirow{2}*{12.7}} \\
                 (Tabletop) & & & & & & & & & & & & & & & & & & & & & & & \\
                \hline
                \textbf{SynTable-Sim} & \multirow{2}*{\textbf{81.0}} & \multirow{2}*{\textbf{82.5}} & \multirow{2}*{\textbf{80.9}} & \multirow{2}*{\textbf{59.1}} & \multirow{2}*{\textbf{66.8}} & \multirow{2}*{\textbf{61.8}} & \multirow{2}*{\textbf{78.1}} & \multirow{2}*{\textbf{69.3}} & \multirow{2}*{\textbf{51.8}} & \multirow{2}*{\textbf{52.4}} & \multirow{2}*{34.6} & \multirow{2}*{\textbf{42.6}} & \multirow{2}*{\textbf{31.2}} & \multirow{2}*{\textbf{41.3}} & \multirow{2}*{\textbf{80.1}} & \multirow{2}*{\textbf{83.2}} & \multirow{2}*{\textbf{81.1}} & \multirow{2}*{\textbf{62.4}} & \multirow{2}*{\textbf{68.1}} & \multirow{2}*{\textbf{64.3}} & \multirow{2}*{\textbf{74.4}} & \multirow{2}*{\textbf{75.7}} & \multirow{2}*{\textbf{86.7}} & \multicolumn{1}{c}{\multirow{2}*{\textbf{82.9}}} \\
                \textbf{(Ours)} & & & & & & & & & & & & & & & & & & & & & & & \\
                \hline
                \hline
            \end{tabular}
        \end{center}
\vspace{-4mm}
\end{table*}

\begin{table*}[!ht]
        \scriptsize
	\captionsetup{justification=centering}
	\renewcommand\arraystretch{1.0}
        \setlength\tabcolsep{7.5pt}
	\begin{center}
	\caption{The performance of UOAIS-Net on the \textbf{SynTable-Sim validation} dataset after training on the UOAIS-Sim and SynTable-Sim datasets. UOAIS-Net is trained with RGB-D images. \textbf{OV}: Overlap F-measure, \textbf{BO}: Boundary F-measure, \textbf{F@.75}: Percentage of segmented objects with an Overlap F-measure greater than 0.75, \textbf{$\textit{F}_\mathcal{O}$}: Occlusion F-Measure, \textbf{$\text{ACC}_{\text{OO}}$}: Occlusion Order Accuracy}	
		\label{eval_on_syntable_validation}
		\begin{tabular}{p{2.7cm}<{\centering}|p{0.6cm}<{\centering}|p{0.6cm}<{\centering}|p{0.7cm}<{\centering}|p{0.6cm}<{\centering}|p{0.6cm}<{\centering}|p{0.7cm}<{\centering}|p{0.6cm}<{\centering}|p{0.6cm}<{\centering}|p{0.6cm}<{\centering}|p{0.6cm}<{\centering}|p{0.7cm}<{\centering}|p{0.8cm}<{\centering}}
                \hline
                \hline
                \multirow{2}*{Training Set} & \multicolumn{3}{c|}{Amodal Mask} & \multicolumn{3}{c|}{Invisible Mask} & \multicolumn{2}{c|}{Occlusion} & \multicolumn{3}{c|}{Visible Mask} & \multirow{2}*{$ACC_{OO}$} \\
                & \multicolumn{1}{c}{OV} & \multicolumn{1}{c}{BO} & \textit{F@.75} & \multicolumn{1}{c}{OV} & \multicolumn{1}{c}{BO} & \textit{F@.75} & \multicolumn{1}{c}{$\textit{F}_\mathcal{O}$} & $\textit{ACC}_\mathcal{O}$ & \multicolumn{1}{c}{OV} & \multicolumn{1}{c}{BO} & \textit{F@.75} & \\
                \hline
                \multirow{1}*{UOAIS-Sim (Tabletop)} & 38.0 & 37.8 & 35.9 & 14.1 & 12.9 & 7.6 & 47.2 & 72.9 & 40.4 & 38.9 & 34.8 & 31.6 \\
                \hline
                \multirow{1}*{\textbf{SynTable-Sim (Ours)}} & \textbf{84.5} & \textbf{78.4} & \textbf{75.6} & \textbf{41.4} & \textbf{37.7} & \textbf{21.5} & \textbf{76.1} & \textbf{82.4} & \textbf{86.8} & \textbf{81.8} & \textbf{74.4} & \textbf{77.5} \\
                \hline
                \hline
            \end{tabular}
        \end{center}
\vspace{-4mm}
\end{table*}

Table \ref{eval_on_OSD_amodal} compares the performance of UOAIS-Net on the OSD-Amodal dataset after training on the UOAIS-Sim tabletop dataset and our SynTable-Sim sample dataset. We conducted four sets of experiments. In each set of experiments, we vary the amount of data augmentation used and the size of the dataset we use for training. 

In our first set of experiments, we can see that the UOAIS-Net trained on the SynTable-Sim dataset significantly outperforms the UOAIS-Net trained on the UOAIS-Sim tabletop dataset in all metrics. Even when we train UOAIS-Net using a dataset of the same size as UOAIS-Sim (SynTable-Sim-0.5X), the performance is still remarkably better than the UOAIS-Net trained on the UOAIS-Sim tabletop dataset across all metrics.  A detailed breakdown of the precision P, recall R, and F-measure F, and \textit{F@.75} scores for the amodal, invisible and visible masks for our first set of experiments is shown in Table \ref{comparing_PRF_on_OSD_amodal_1}. We observe that except for the Boundary precision scores of the invisible masks, UOAIS-Net achieves substantial improvements in all other metrics. 

In the next three sets of experiments, we observe that even when we include data augmentation, the performance of UOAIS-Net trained on the UOAIS-Sim tabletop dataset is still worse than that trained on the SynTable-Sim dataset without using any data augmentation.  We also provide images of the inference results on the OSD-Amodal dataset in Section \ref{sec6} of our supplementary materials.

Similarly, from Table \ref{eval_on_syntable_validation}, the UOAIS-Net model trained on the SynTable-Sim dataset outperforms the one trained on UOAIS-Sim tabletop dataset in all metrics when both models are benchmarked on SynTable-Sim validation dataset. 



We evaluated the effectiveness of SynTable-Sim across different UOAIS models comprising distinct architectures. Table \ref{eval_on_OSD_Amodal_all_models} compares the performance of UOAIS models--- Amodal MRCNN, ORCNN, ASN, and UOAIS-Net---on the OSD-Amodal dataset after training on the UOAIS-Sim tabletop dataset and our SynTable-Sim sample dataset. For each model result in our experiments, we used seed 7 for training. Generally, across most metrics, the UOAIS models trained on SynTable-Sim outperform the same models trained on the UOAIS-Sim tabletop dataset. There is also a significant improvement in the results of $ACC_{oo}$ for Amodal MRCNN, ORCNN, and ASN when trained on our SynTable-Sim as compared to the UOAIS-Sim tabletop dataset. This is consistent with the performance trend observed for UOAIS-Net and, therefore, demonstrates that SynTable is an effective tool for generating high-quality datasets that can improve the performance of UOAIS models. A detailed breakdown of the precision P, recall R, and F-measure F, and \textit{F@.75} scores for the amodal, invisible, and visible masks are shown in Table \ref{eval_on_OSD_Amodal_all_models_breakdown}.

As shown in Table \ref{eval_on_syntable_validation_all_models}, the UOAIS models trained on the SynTable-Sim dataset outperform the same models trained on the UOAIS-Sim tabletop dataset in all metrics when they are benchmarked on the SynTable-Sim validation dataset. 

\begin{table*}[!ht]
        \footnotesize
	\captionsetup{justification=centering}
	\renewcommand\arraystretch{1.0}
        \setlength\tabcolsep{5pt}
	\begin{center}
	\caption{The performance of Amodal MRCNN, ORCNN, ASN, and UOAIS-Net on the \textbf{OSD-Amodal dataset} after training on the UOAIS-Sim and SynTable-Sim datasets. UOAIS-Net is trained with RGB-D images. \textbf{OV}: Overlap F-measure, \textbf{BO}: Boundary F-measure, \textbf{F@.75}: Percentage of segmented objects with an Overlap F-measure greater than 0.75, \textbf{$\text{ACC}_{\text{OO}}$}: Occlusion Order Accuracy}	
		\label{eval_on_OSD_Amodal_all_models}
		\begin{tabular}{p{2.5cm}<{\centering}|p{2.1cm}<{\centering}|p{0.5cm}<{\centering}|p{0.5cm}<{\centering}|p{0.7cm}<{\centering}|p{0.5cm}<{\centering}|p{0.5cm}<{\centering}|p{0.7cm}<{\centering}|p{0.5cm}<{\centering}|p{0.6cm}<{\centering}|p{0.5cm}<{\centering}|p{0.5cm}<{\centering}|p{0.7cm}<{\centering}|p{0.7cm}<{\centering}}
                \hline
                \hline
                \multirow{2}*{Training Set} & \multirow{2}*{Method} & \multicolumn{3}{c|}{Amodal Mask} & \multicolumn{3}{c|}{Invisible Mask} & \multicolumn{2}{c|}{Occlusion} & \multicolumn{3}{c|}{Visible Mask} & \multirow{2}*{$ACC_{OO}$} \\
                & & \multicolumn{1}{c}{OV} & \multicolumn{1}{c}{BO} & \textit{F@.75} & \multicolumn{1}{c}{OV} & \multicolumn{1}{c}{BO} & \textit{F@.75} & \multicolumn{1}{c}{$\textit{F}_\mathcal{O}$} & $\textit{ACC}_\mathcal{O}$ & \multicolumn{1}{c}{OV} & \multicolumn{1}{c}{BO} & \textit{F@.75} & \\
                \hline
                \multirow{4}*{UOAIS-Sim (Tabletop)} & Amodal MRCNN & 36.7 & 26.9 & 45.7 & 8.8 & 4.8 & 7.7 & 39.2 & 54.8 & 38.7 & 26.3 & 32.2 & 15.6 \\ 
                & ORCNN & 36.3 & 25.4 & 47.0 & 12.2 & 6.7 & 9.0 & 43.8 & 59.2 & 30.5 & 21.8 & 29.6 & 21.5 \\ 
                & ASN & 40.5 & 33.6 & 49.8 & 17.4 & 12.1 & 15.0 & 47.0 & 63.2 & 39.3 & 31.6 & 36.8 & 17.8 \\ 
                & UOAIS-Net & 49.0 & 50.3 & 82.7 & 42.3 & 23.9 & 40.3 & 68.9 & 84.0 & 47.3 & 50.0 & 70.6 & 80.4 \\
                \hline
                \multirow{4}*{\textbf{SynTable-Sim (Ours)}} & Amodal MRCNN & 74.5 & 57.5 & 77.2 & 41.3 & 23.5 & 37.6 & 69.3 & 79.4 & 73.8 & 57.7 & 66.1 & 79.2 \\ 
                & ORCNN & 74.2 & 58.2 & 77.1 & 44.7 & 24.3 & 33.8 & 72.9 & 82.2 & 72.0 & 58.3 & 67.7 & 79.1 \\ 
                & ASN & 78.2 & 60.2 & 75.3 & 46.4 & 27.7 & 35.8 & 72.6 & 83.0 & 78.1 & 61.8 & 68.9 & 80.2 \\ 
                & UOAIS-Net & 64.4 & 51.5 & 84.3 & 47.3 & 24.2 & 47.4 & 60.0 & 91.9 & 65.3 & 53.7 & 78.2 & 87.0 \\
                \hline
                \hline
            \end{tabular}
        \end{center}
\end{table*}

\begin{table*}[!ht]
        \footnotesize
	\captionsetup{justification=centering}
	\renewcommand\arraystretch{1.0}
        \setlength\tabcolsep{4pt}
	\begin{center}
		\caption{A breakdown of the precision, recall, and F-measure of the amodal, invisible, and visible mask predictions by Amodal MRCNN, ORCNN, ASN, and UOAIS-Net on the \textbf{OSD-Amodal dataset} after training on the UOAIS-Sim and SynTable-Sim dataset. \textbf{P}: Precision, \textbf{R}: Recall, \textbf{F}: F-measure}
		\label{eval_on_OSD_Amodal_all_models_breakdown}
        \setlength\tabcolsep{1.5pt}
		\begin{tabular}{p{1.7cm}<{\centering}|p{1.4cm}<{\centering}|
                            p{0.55cm}<{\centering}|p{0.55cm}<{\centering}|p{0.55cm}<{\centering}|
                            p{0.55cm}<{\centering}|p{0.55cm}<{\centering}|p{0.55cm}<{\centering}|p{0.75cm}<{\centering}|
                            p{0.55cm}<{\centering}|p{0.55cm}<{\centering}|p{0.55cm}<{\centering}|
                            p{0.55cm}<{\centering}|p{0.55cm}<{\centering}|p{0.55cm}<{\centering}|p{0.75cm}<{\centering}|
                            p{0.55cm}<{\centering}|p{0.55cm}<{\centering}|p{0.55cm}<{\centering}|
                            p{0.55cm}<{\centering}|p{0.55cm}<{\centering}|p{0.55cm}<{\centering}|p{0.75cm}<{\centering}}
                \hline
                \hline
                \multirow{3}*{Training Set} & \multirow{3}*{Method} & \multicolumn{7}{c|}{Amodal Mask} & \multicolumn{7}{c|}{Invisible Mask}& \multicolumn{7}{c}{Visible Mask} \\
                \cline{3-23}
                & & \multicolumn{3}{c|}{Overlap} & \multicolumn{3}{c|}{Boundary} & \multirow{2}*{\textit{F@.75}} & \multicolumn{3}{c|}{Overlap} & \multicolumn{3}{c|}{Boundary} & \multirow{2}*{\textit{F@.75}}
                 & \multicolumn{3}{c|}{Overlap} & \multicolumn{3}{c|}{Boundary} & \multirow{2}*{\textit{F@.75}} \\
                & & \multicolumn{1}{c}{P} & \multicolumn{1}{c}{R} & F & \multicolumn{1}{c}{P} & \multicolumn{1}{c}{R} & F & & \multicolumn{1}{c}{P} & \multicolumn{1}{c}{R} & F & \multicolumn{1}{c}{P} & \multicolumn{1}{c}{R} & F & & \multicolumn{1}{c}{P} & \multicolumn{1}{c}{R} & F & \multicolumn{1}{c}{P} & \multicolumn{1}{c}{R} & F &\\
                \hline
                 \multirowcell{5}{UOAIS-Sim\\(Tabletop)} & Amodal & \multirow{2}*{27.9} &  \multirow{2}*{66.7} & \multirow{2}*{36.7} &  \multirow{2}*{22.5} &  \multirow{2}*{39.8} &  \multirow{2}*{26.9} &  \multirow{2}*{45.7} & \multirow{2}*{20.2} &  \multirow{2}*{24.9} &  \multirow{2}*{8.8} &  \multirow{2}*{16.4} &  \multirow{2}*{19.9} & \multirow{2}*{4.8} &  \multirow{2}*{7.7} &  \multirow{2}*{30.1} & \multirow{2}*{60.5} & \multirow{2}*{38.7} &  \multirow{2}*{22.0} & \multirow{2}*{37.8} & \multirow{2}*{26.3} & \multirow{2}*{32.2}\\
                 & MRCNN & & & & & & & & & & & & & & & & & & & & &\\
                 & ORCNN & 26.3 & 71.1 & 36.3 & 19.8 & 42.4 & 25.4 & 47.0 & 41.5 & 22.7 & 12.2 & 33.9 & 17.9 & 6.7 & 9.0 & 21.4 & 63.4 & 30.5 & 16.6 & 38.2 & 21.8 & 29.6 \\ 
                 & ASN & 31.7 & 67.8 & 40.5 & 28.6 & 45.4 & 33.6 & 49.8 & 47.6 & 23.4 & 17.4 & 38.8 & 20.2 & 12.1 & 15.0 & 32.0 & 63.5 & 39.3 & 28.2 & 41.5 & 31.6 & 36.8\\ 
                 & UOAIS-Net & 37.2 &  85.5 & 49.0 & 41.1 & 71.3 & 50.3 & 82.7 &  50.9 &  54.0 & 42.3 & 24.8 & 41.1 & 23.9 & 40.3 & 35.4 & 81.6 & 47.3 &  41.3 & 69.3 & 50.0 & 70.6\\
                \hline
                \multirowcell{5}{\textbf{SynTable-Sim}\\\textbf{(Ours)}} & Amodal & \multirow{2}*{72.3} &  \multirow{2}*{81.6} & \multirow{2}*{74.5} &  \multirow{2}*{54.6} &  \multirow{2}*{64.5} &  \multirow{2}*{57.5} &  \multirow{2}*{77.2} & \multirow{2}*{54.9} &  \multirow{2}*{48.0} &  \multirow{2}*{41.3} &  \multirow{2}*{30.3} &  \multirow{2}*{38.8} &  \multirow{2}*{23.5} & \multirow{2}*{37.6} &  \multirow{2}*{72.1} & \multirow{2}*{78.4} & \multirow{2}*{73.8} &  \multirow{2}*{55.1} & \multirow{2}*{63.7} & \multirow{2}*{57.7} & \multirow{2}*{66.1}\\
                 & MRCNN & & & & & & & & & & & & & & & & & & & & &\\
                 & ORCNN & 73.7 & 80.8 & 74.2 & 55.6 & 64.5 & 58.2 & 77.1 & 61.0 & 47.1 & 44.7 & 31.1 & 38.6 & 24.3 & 33.8 & 69.8 & 79.1 & 72.0 & 55.7 & 64.3 & 58.3 & 67.7 \\ 
                 & ASN & 78.2 & 80.3 & 78.2 & 57.8 & 64.9 & 60.2 & 75.3 & 65.2 & 46.2 & 46.4 & 32.9 & 38.7 & 27.7 & 35.8 & 77.5 & 80.1 & 78.1 & 60.4 & 65.4 & 61.8 & 68.9\\ 
                 & UOAIS-Net & 53.9 & 86.3 & 64.4 & 40.9 & 74.6 & 51.5 & 84.3 & 53.0 & 60.0 & 47.3 & 20.5 & 48.3 & 24.2 & 47.4 & 55.0 & 86.2 & 65.3 & 43.2 & 75.3 & 53.7 & 78.2\\
                \hline
                \hline
            \end{tabular}
        \end{center}
\end{table*}

\begin{table*}[!t]
        \footnotesize
	\captionsetup{justification=centering}
	\renewcommand\arraystretch{1.0}
        \setlength\tabcolsep{5pt}
	\begin{center}
	\caption{The performance of Amodal MRCNN, ORCNN, ASN, and UOAIS-Net on the \textbf{SynTable-Sim validation dataset} after training on the UOAIS-Sim and SynTable-Sim datasets. UOAIS-Net is trained with RGB-D images. \textbf{OV}: Overlap F-measure, \textbf{BO}: Boundary F-measure, \textbf{F@.75}: Percentage of segmented objects with an Overlap F-measure greater than 0.75, \textbf{$\text{ACC}_{\text{OO}}$}: Occlusion Order Accuracy}	
		\label{eval_on_syntable_validation_all_models}
		\begin{tabular}{p{2.5cm}<{\centering}|p{2.1cm}<{\centering}|p{0.5cm}<{\centering}|p{0.5cm}<{\centering}|p{0.7cm}<{\centering}|p{0.5cm}<{\centering}|p{0.5cm}<{\centering}|p{0.7cm}<{\centering}|p{0.5cm}<{\centering}|p{0.6cm}<{\centering}|p{0.5cm}<{\centering}|p{0.5cm}<{\centering}|p{0.7cm}<{\centering}|p{0.7cm}<{\centering}}
                \hline
                \hline
                \multirow{2}*{Training Set} & \multirow{2}*{Method} & \multicolumn{3}{c|}{Amodal Mask} & \multicolumn{3}{c|}{Invisible Mask} & \multicolumn{2}{c|}{Occlusion} & \multicolumn{3}{c|}{Visible Mask} & \multirow{2}*{$ACC_{OO}$} \\
                & & \multicolumn{1}{c}{OV} & \multicolumn{1}{c}{BO} & \textit{F@.75} & \multicolumn{1}{c}{OV} & \multicolumn{1}{c}{BO} & \textit{F@.75} & \multicolumn{1}{c}{$\textit{F}_\mathcal{O}$} & $\textit{ACC}_\mathcal{O}$ & \multicolumn{1}{c}{OV} & \multicolumn{1}{c}{BO} & \textit{F@.75} & \\
                \hline
                \multirow{4}*{UOAIS-Sim (Tabletop)} & Amodal MRCNN & 27.1 & 25.2 & 23.8 & 6.7 & 6.2 & 3.5 & 35.8 & 66.0 & 29.1 & 26.2 & 23.5 & 19.0 \\ 
                & ORCNN & 30.9 & 29.0 & 28.1 & 12.5 & 11.4 & 8.0 & 39.9 & 68.4 & 31.8 & 30.2 & 27.3 & 23.1 \\ 
                & ASN & 33.3 & 34.4 & 35.3 & 10.3 & 9.1 & 5.0 & 47.6 & 72.3 & 35.0 & 36.0 & 34.1 & 31.6 \\ 
                & UOAIS-Net & 39.9 & 40.5 & 38.6 & 17.0 & 15.5 & 9.6 &  49.6 & 74.9 & 41.6 & 40.7 & 35.9 & 31.6 \\
                \hline
                \multirow{4}*{\textbf{SynTable-Sim (Ours)}} & Amodal MRCNN & 83.5 & 76.2 & 72.5 & 35.4 & 31.8 & 16.4 & 73.2 & 80.3 & 85.7 & 79.1 & 71.1 & 72.8 \\ 
                & ORCNN & 83.4 & 76.0 & 72.2 & 34.4 & 29.3 & 15.3 & 67.2 & 73.7 & 85.3 & 78.9 & 70.9 & 73.0 \\ 
                & ASN & 83.6 & 76.9 & 73.9 & 38.5 & 35.1 & 18.5 & 74.8 & 81.5 & 86.1 & 80.0 & 72.8 & 75.8 \\ 
                & UOAIS-Net & 83.7 & 77.5 & 75.1 & 40.3 & 36.7 & 20.2 & 75.5 & 82.0 & 86.2 & 80.1 & 73.3 & 77.4 \\
                \hline
                \hline
            \end{tabular}
        \end{center}
\end{table*}

Our experiments demonstrate the effectiveness of our proposed dataset generation pipeline, SynTable, in improving the Sim-to-Real transfer performance of SOTA deep learning computer vision models for UOAIS. These results highlight the potential of SynTable for addressing the challenge of annotating amodal instance segmentation masks.

\vspace{-2mm}
\section{Conclusion} 
\label{sec:conclusion}

In conclusion, we present SynTable, a novel synthetic data generation pipeline for generating photorealistic datasets that facilitated amodal instance segmentation of cluttered tabletop scenes. SynTable enables the creation of complex 3D scenes with automatic annotation of diverse metadata, eliminating the need for manual labeling while ensuring dataset quality and accuracy. We demonstrate the effectiveness of the SynTable pipeline by generating a photorealistic amodal instance segmentation dataset and using it to train UOAIS-Net. As a result, UOAIS-Net achieves significantly improved Sim-to-Real transfer performance on the OSD-Amodal dataset, particularly in determining the object occlusion order of objects in a cluttered tabletop scene. SynTable advances amodal segmentation for systems that require occlusion-aware perception, such as robotics, augmented reality. By automating annotation of amodal masks and appearance via photorealistic rendering, and scene occlusion order, our pipeline addresses a key bottleneck in training robust vision models with amodal perception capabilities.

{
    \small
    \bibliographystyle{ieeenat_fullname}
    \bibliography{main}
}

\clearpage
\setcounter{page}{1}
\maketitlesupplementary

\section{Overview}
This supplementary material offers dataset visualization, qualitative results, and additional technical details to support the main paper. Section \ref{sec:supp_method} provides additional information about each step in our dataset generation process. Section \ref{sec3} provides a comprehensive elaboration of the evaluation metrics employed. Section \ref{sec4} illustrates how occlusion order accuracy is calculated and the validity of the metric. Furthermore, Section \ref{sec5} delineates the process of generating an occlusion order directed acyclic graph from the occlusion order adjacency matrix to classify objects in three distinct order layers. Lastly, Section \ref{sec6} showcases some qualitative inference results of UOAIS-Net on the OSD-Amodal dataset. 


\subsection{Video Demonstration of SynTable-Sim Generation Process}
In addition to this document, we include a demonstration video as part of our supplementary material to demonstrate in detail the process of generating a custom synthetic dataset using SynTable. We refer readers to the demonstration video for a detailed visualization of the dataset generation process. The video can be found at \url{https://www.youtube.com/watch?v=zHM8H58Kn3E}.

\subsection{Management of the SynTable-Sim Dataset}
The source code for our work is available at \url{https://github.com/ngzhili/SynTable}. All the CAD models of the objects used in our SynTable-Sim dataset, as well as the dataset itself, are hosted in the Zenodo open repository, free for all to download. The DOI of our dataset is \url{10.5281/zenodo.10565517}. The dataset can be accessed at \url{https://doi.org/10.5281/zenodo.10565517}


\section{Additional Details About the Dataset Generation Process}
\label{sec:supp_method}



\subsection{Preparing Each Scene}
\label{sec:scene_preparation}
\begin{figure}[t]
	\centering
	\includegraphics[scale=0.65]{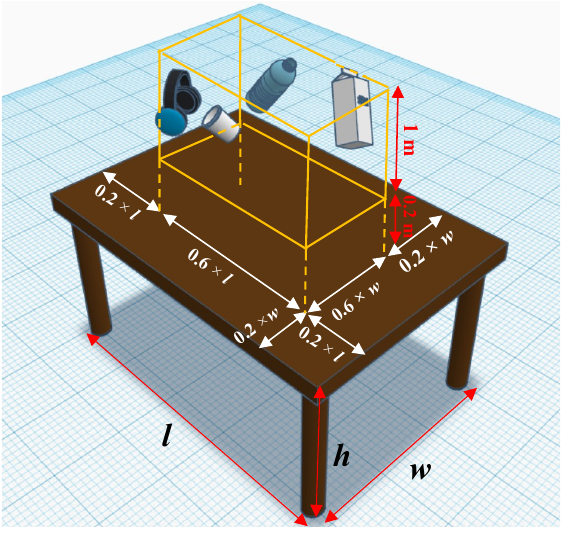}
	\caption{Initialization of objects with randomized coordinates and rotations. The initial position of the objects in the scene is randomized but constrained to be within the dimensions of the 3D orange box. The orange box is 0.2 m above the tabletop. The roll, pitch, and yaw of each object are also randomly sampled within the range of $0\degree$ to $360\degree$.}
	\label{prepare_scene}
\vspace{-4mm}
\end{figure}

The method to prepare each scene is shown in Figure \ref{prepare_scene}. A table is randomly sampled from the assets in Omniverse Nucleus and is rendered at the center of a room. The texture and materials of the table, ceiling, wall, and floor are randomized for every scene to ensure domain randomization. The objects are added to the scene with randomized $x$, $y$, and $z$ coordinates and orientations. We randomly sample (with replacement) $N_{lower}$ to $N_{upper}$ objects to render for each scene. By default, $N_{lower}=1$, $N_{upper}=40$. Each object is initialized with real-life dimensions, randomized rotations and coordinates, allowing for diverse object arrangements across scenes. Each object also has mass and collision properties so that they can be dropped onto the tabletop in our physics simulation.

\subsection{Physical Simulation of Each Scene}
\label{sec:physical_simulation}
Upon completing the scene preparation, the rendered objects are dropped onto the table surface using a physics simulation. The simulation is paused after $t$ seconds ($t=5$ by default), halting any further movement of the objects. During the simulation, any objects that rebound off the tabletop surface and fall outside the spatial coordinate region of the tabletop surface (i.e., either below the table or beyond the width and length of the table) are automatically removed. This is necessary to prevent the inclusion of extraneous and irrelevant objects outside the specified tabletop region during the annotation process from different viewpoints.

\subsection{Sampling of Camera Viewpoints}
\label{sec:sample_camera_viewpoint}
\begin{figure}[t]
	\centering
	\includegraphics[width=\linewidth]{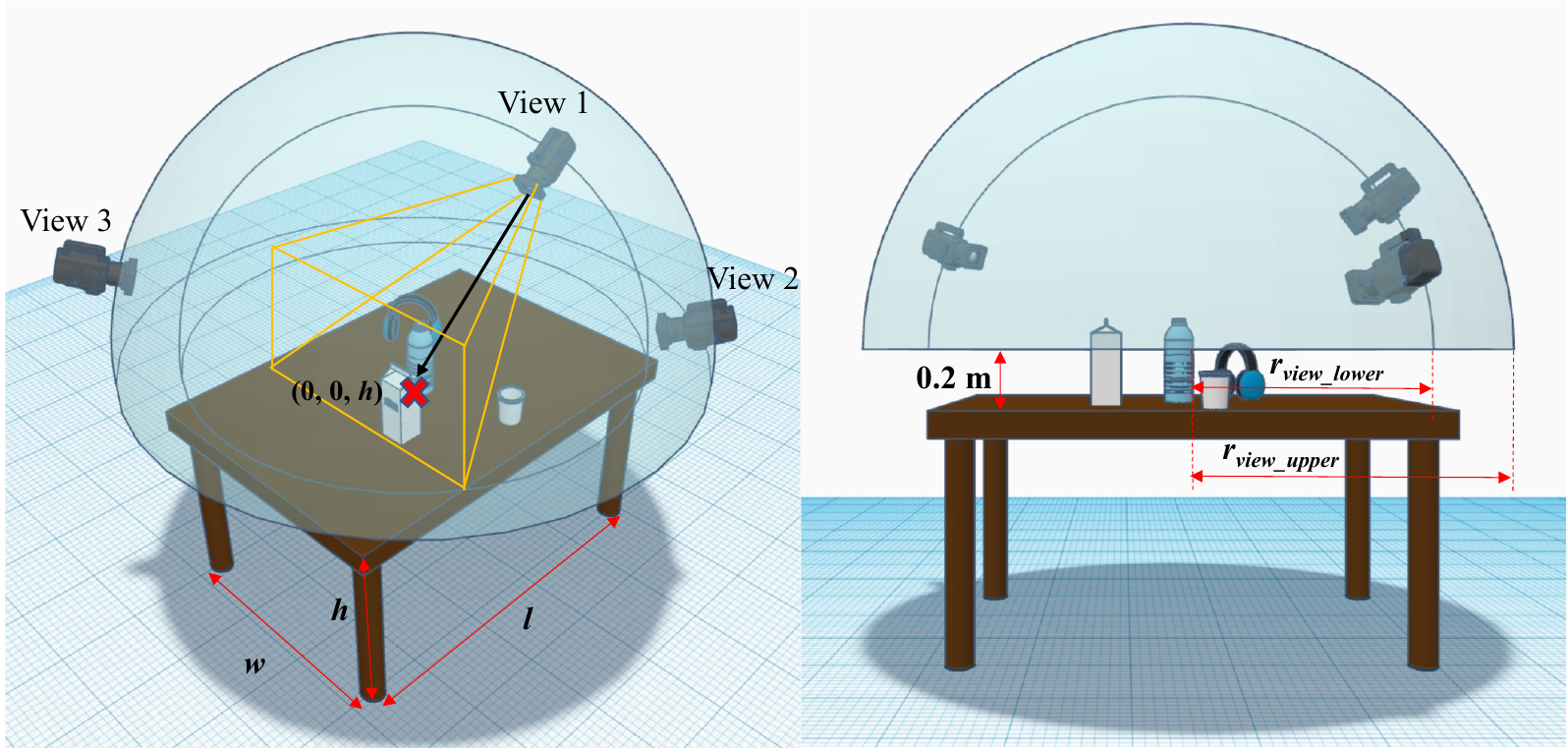}	
        \caption{Sampling of camera viewpoints within concentric hemispheres (shown in blue). The two concentric hemispheres’ origins are centered at the tabletop surface’s center coordinate with an offset of 0.2 m in the positive $z$ direction in the world frame. This allows the camera viewpoints to minimally have a direct line of sight to the tabletop surface to capture part of the tabletop plane. This figure is best viewed zoomed in.}
	\label{camera_viewpoints}
\vspace{-4mm}
\end{figure}

To capture annotations for each scene from multiple viewpoints, we enhance the approach by Gilles \textit{et al.} \cite{metagraspnet2022}---which only uses fixed viewpoint positions---by introducing a feature that captures $V$ number of viewpoints at random positions within two concentric hemispheres, as illustrated in Figure \ref{camera_viewpoints}. $V$ can be set by the user. The radii of the two concentric hemispheres are uniformly sampled within the range $r_{view\_lower}$ m to $r_{view\_upper}$ m, where $r_{view\_lower}$ and $r_{view\_upper}$ are defined in Equations \ref{eqn:hemi_view_lower} and \ref{eqn:hemi_view_upper}. Users may also set fixed values for $r_{view\_lower}$ and $r_{view\_upper}$ should they wish to do so.

\begin{equation}
r_{view\_lower} = max\left(\frac{w}{2}, \frac{l}{2}\right)
\label{eqn:hemi_view_lower}
\end{equation}
\begin{equation}
r_{view\_upper} = 1.7 \times r_{view\_lower}
\label{eqn:hemi_view_upper}
\end{equation}

The hemisphere’s spherical coordinates are parameterized using three variables $r_{view}$, $u$, and $v$. To generate the camera coordinates in the world frame, we first obtain the radius of the hemisphere $r_{view}$ by uniform sampling between $r_{view\_lower}$ and  $r_{view\_upper}$. Next, we uniformly sample $u,v$ $\in$ [0,1], then substitute all the sampled values into Equations \ref{eqn:hemi3}, \ref{eqn:hemi4} and \ref{eqn:hemi5} to compute the cartesian coordinates of the camera.

\begin{equation}
x = r_{view} \sin(\arccos(1-v))\cos(2 \pi u)
\label{eqn:hemi3}
\end{equation}
\begin{equation}
y = r_{view} \sin(\arccos(1-v))\sin(2 \pi u)
\label{eqn:hemi4}
\end{equation}
\begin{equation}
z = r_{view} \cos(\arccos(1-v))
\label{eqn:hemi5}
\end{equation}

Once the camera coordinates are set, the orientation of each camera is set such that each viewpoint looks directly at the center of the tabletop surface (0, 0, $h$).

\subsection{Sampling of Lighting Conditions}
\label{supp_sec:sample_lighting}

\begin{figure}[t]
	\centering
	\includegraphics[scale=0.45]{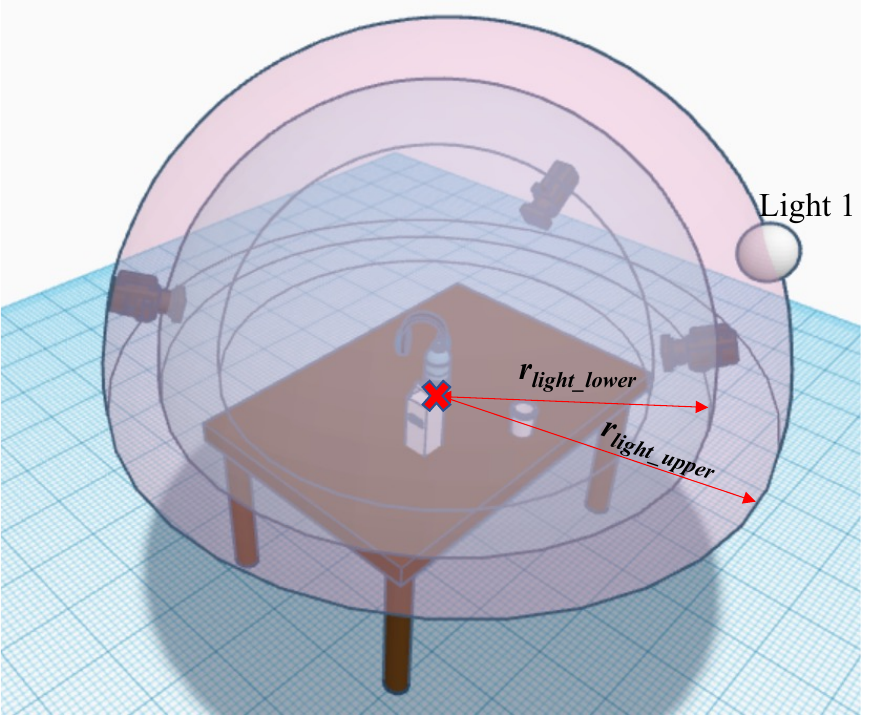}
	\caption{Sampling of lighting within concentric hemispheres (shown in pink). Each spherical light source lies within the constraints of two concentric hemispheres of arbitrary radius between $r_{light\_lower}$ to $r_{light\_upper}$. Note that the radii constraints for the spherical light source concentric hemispheres are larger than those for the camera viewpoints' and are customizable by the user.}
	\label{lighting}	
\vspace{-4mm}
\end{figure}

To simulate different indoor lighting conditions, we resample $L$ spherical light sources between $L_{lower}$ to $L_{upper}$ for each viewpoint (Figure \ref{lighting}). By default, we set $L_{lower}$ and $L_{upper}$ to be 0 and 2, respectively. To position $L$ spherical light sources for a viewpoint, we adopt a similar approach to the camera viewpoint sampling method discussed in Section \ref{sec:sample_camera_viewpoint}. In contrast to the approach by Back \textit{et al.} \cite{back2022unseen}, we use spherical light sources that emit light in all directions. Furthermore, we uniformly sample light source temperatures between 2,000 K to 6,500 K. The default light intensity of each light source is uniformly sampled between 100 lx to 20,000 lx, and the default light intensity of ceiling lights in the scene is also sampled uniformly between 100 lx to 2,000 lx. To achieve diverse indoor lighting conditions for tabletop scenes, users have the flexibility to adjust the number of spherical light sources, as well as their intensities and temperatures.

Similar to the sampling method for the camera viewpoint coordinates, we have designed a feature that samples the lower and upper radii bounds for the light sources based on the camera hemisphere's upper bound radius, $r_{view\_upper}$. The sampled lower and upper bound radii constraints for the lighting hemisphere $r_{light\_lower}$ and $r_{light\_upper}$ are as follows:

\vspace{-2mm}
\begin{equation}
r_{light\_lower} = r_{view\_upper} + 0.1 m
\label{eqn:hemi_light_lower}
\end{equation}
\begin{equation}
r_{light\_upper} = r_{light\_lower} + 1 m
\label{eqn:hemi_light_upper}
\vspace{-2mm}
\end{equation}

\subsection{Saving of Ground Truth Annotations}
We saved the RGB and depth images as PNG images. The OOAM of the objects in each image is saved as a NumPy file. The amodal, visible, and occlusion masks are saved as Run-length Encoding (RLE) in COCO JSON format to optimize disk space used by the generated datasets. We also recorded each object’s visible bounding box, image ID, and object name in the generated COCO JSON file.


\begin{figure*}[!ht]
	\centering
	\includegraphics[width=\linewidth]{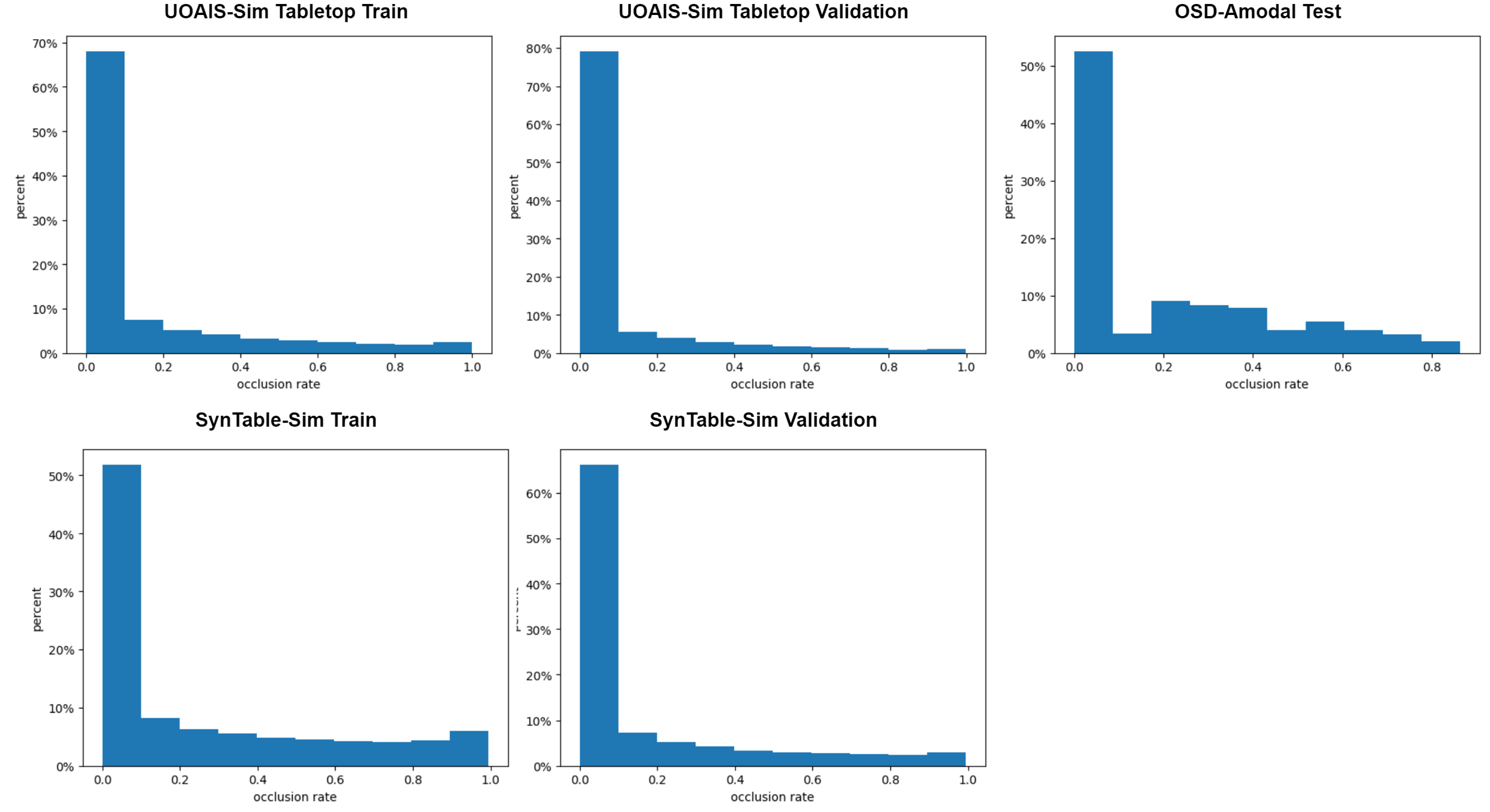}
	\caption{Histogram of occlusion rate for UOAIS-Sim tabletop, SynTable-Sim and OSD-Amodal datasets}
	\label{histogram_occlusion_rate}	
\end{figure*}


\begin{figure*}[!ht]
	\centering
	\includegraphics[width=\linewidth]{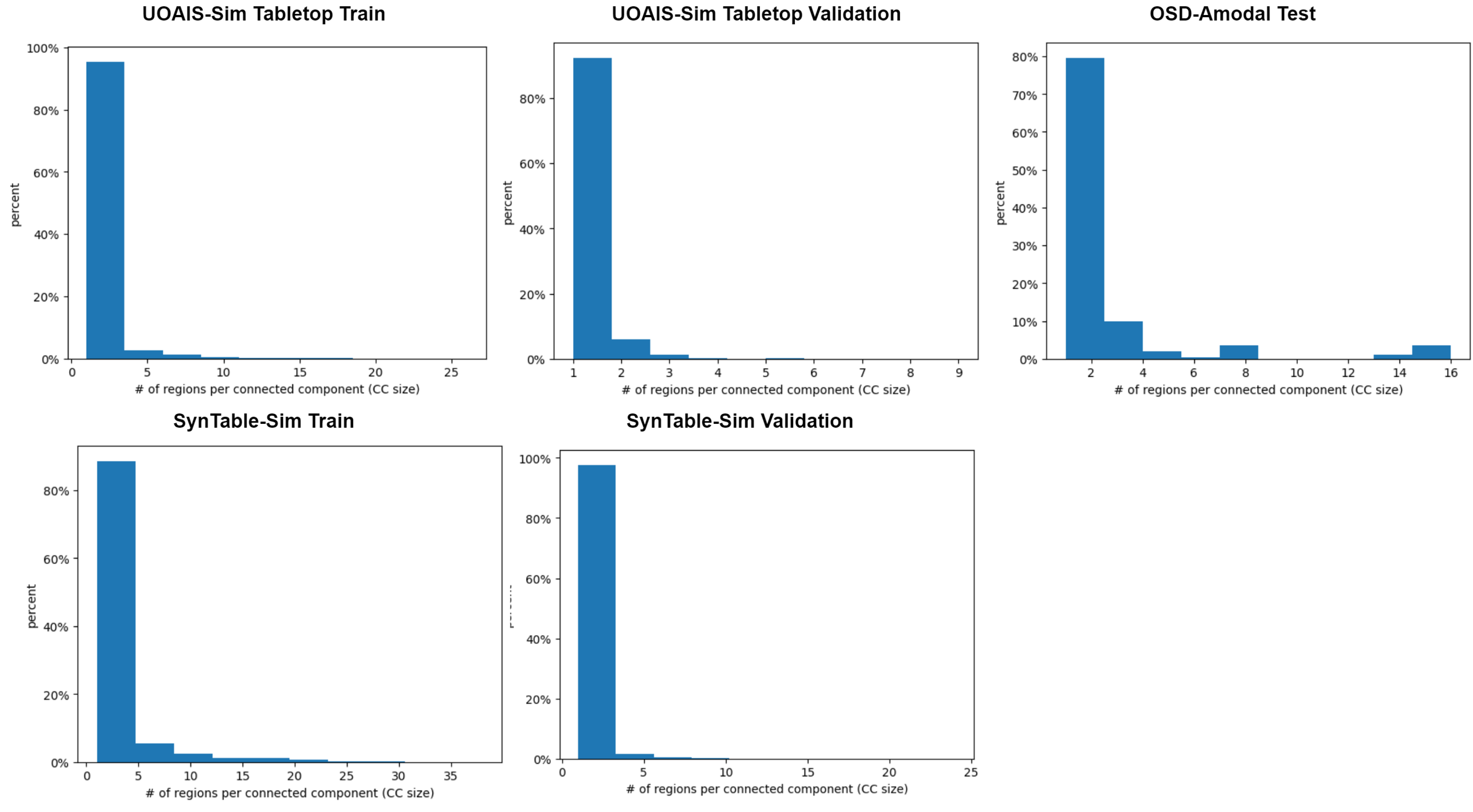}
	\caption{Histogram for number of regions per connected component (connected component size) for UOAIS-Sim tabletop, SynTable-Sim and OSD-Amodal datasets}
	\label{histogram_number_of_regions_per_connected_component}	
\end{figure*}

\section{Details about Evaluation Metrics}
\label{sec3}

In this paper, we employ the precision/recall/F-measure (P/R/F) metrics, as defined in \cite{TOD,Dave2019TowardsSA, segmentation_of_moving_objects}. This metric favors methods that accurately segment the desired objects while penalizing those that produce false positives. Specifically, the precision, recall, and F-measure are calculated between all pairs of predicted and ground truth objects. The Hungarian method, employing pairwise F-measure, is utilized to establish a match between predicted objects and ground truth. Given this matching, the Overlap P/R/F is computed by:

\begin{equation}
P=\frac{\sum_{i}\left|c_{i} \cap g\left(c_{i}\right)\right|}{\sum_{i}\left|c_{i}\right|},\ \ \  R=\frac{\sum_{i}\left|c_{i} \cap g\left(c_{i}\right)\right|}{\sum_{j}\left|g_{j}\right|}
\label{eq1}
\end{equation}

\begin{equation}
F=\frac{2 P R}{P+R}
\label{eq3}
\end{equation}
where \(c_{i}\) denotes the set of pixels belonging to predicted object \(i\), \(g\left(c_{i}\right)\) is the set of pixels of the matched ground truth object of \(c_{i}\) after Hungarian matching, and \(g_{j}\) is the set of pixels for ground truth object \(j\).

Although the aforementioned metric provides valuable information, it fails to consider the boundaries of the objects. Therefore, Xie \etal \cite{TOD} proposed the Boundary P/R/F measure to supplement the Overlap P/R/F. The calculation of Boundary P/R/F involves the same Hungarian matching as used in the computation of Overlap P/R/F. Given these matchings, the Boundary P/R/F is computed by:
\begin{equation}
P=\frac{\sum_{i}\left|c_{i} \cap D\left[g\left(c_{i}\right)\right]\right|}{\sum_{i}\left|c_{i}\right|},\ \ \
R=\frac{\sum_{i}\left|D\left[c_{i}\right] \cap g\left(c_{i}\right)\right|}{\sum_{j}\left|g_{j}\right|}
\label{eq4}
\end{equation}

\begin{equation}
F=\frac{2 P R}{P+R}
\label{eq6}
\end{equation}

Here, overloaded notations are used to represent the sets of pixels belonging to the boundaries of the predicted object $i$ and the ground truth object $j$ as $c_i$ and $g_j$, respectively. The dilation operation is denoted by $D[\cdot]$, which allows for some tolerance in the prediction. The metrics we use are a combination of the F-measure described in \cite{Video_Object_Segmentation} and the Overlap P/R/F as defined in \cite{Dave2019TowardsSA}.


In our work, we use the Overlap and Boundary P/R/F evaluation metrics to evaluate the accuracy of the predicted visible, invisible, and amodal masks. In the context of the Overlap P/R/F metrics, $c_i$ denotes the set of pixels belonging to the predicted visible, invisible, and amodal masks, \(g\left(c_{i}\right)\) denotes the set of pixels belonging to the matched ground-truth visible, invisible and amodal masks annotations, and \(g_{j}\) is the ground-truth visible, invisible and amodal mask. The meaning of $c_i$, \(g\left(c_{i}\right)\), and \(g_{j}\) are similar in the context of the Boundary P/R/F metrics.

An additional vital evaluation metric used in our paper is the \textit{F@.75}. This metric represents the proportion of segmented objects with an Overlap F-measure greater than 0.75. It is important not to confuse this metric with the F-measure computed for the Overlap and Boundary P/R/F. The F-measure for Overlap and Boundary is a harmonic mean of a model's average precision and average recall, while \textit{F@.75} indicates the percentage of objects from a dataset that can be segmented with high accuracy. The \textit{F} in F@.75 refers to the F-measure computed for a ground truth object after the Hungarian matching of the ground truth mask \textit{j} with the predicted mask \textit{i} as defined in \cite{Dave2019TowardsSA} and stated in Equation (\ref{eq8}).

\begin{equation}
P_{i j}=\frac{\left|c_i \cap g_j\right|}{\left|c_i\right|},\ \ \ R_{i j}=\frac{\left|c_i \cap g_j\right|}{\left|g_j\right|}
\label{eq7}
\end{equation}

\begin{equation}
F_{i j}=\frac{2 P_{i j} R_{i j}}{P_{i j}+R_{i j}}
\label{eq8}
\end{equation}

The notation ${c_i}$ denotes the set of pixels that belong to a predicted region $i$, while $g_j$ represents all the pixels that belong to a non-background ground truth region $j$. In addition, $P_{i j}$ represents the precision score, $R_{i j}$ represents the recall score, and $F_{i j}$ represents the F-measure score that corresponds to this particular pair of predicted and ground truth regions.

\section{Occlusion Order Accuracy $ACC_{oo}$ metric}
\label{sec4}
\begin{figure*}[!ht]
	\centering
	\includegraphics[width=\linewidth]{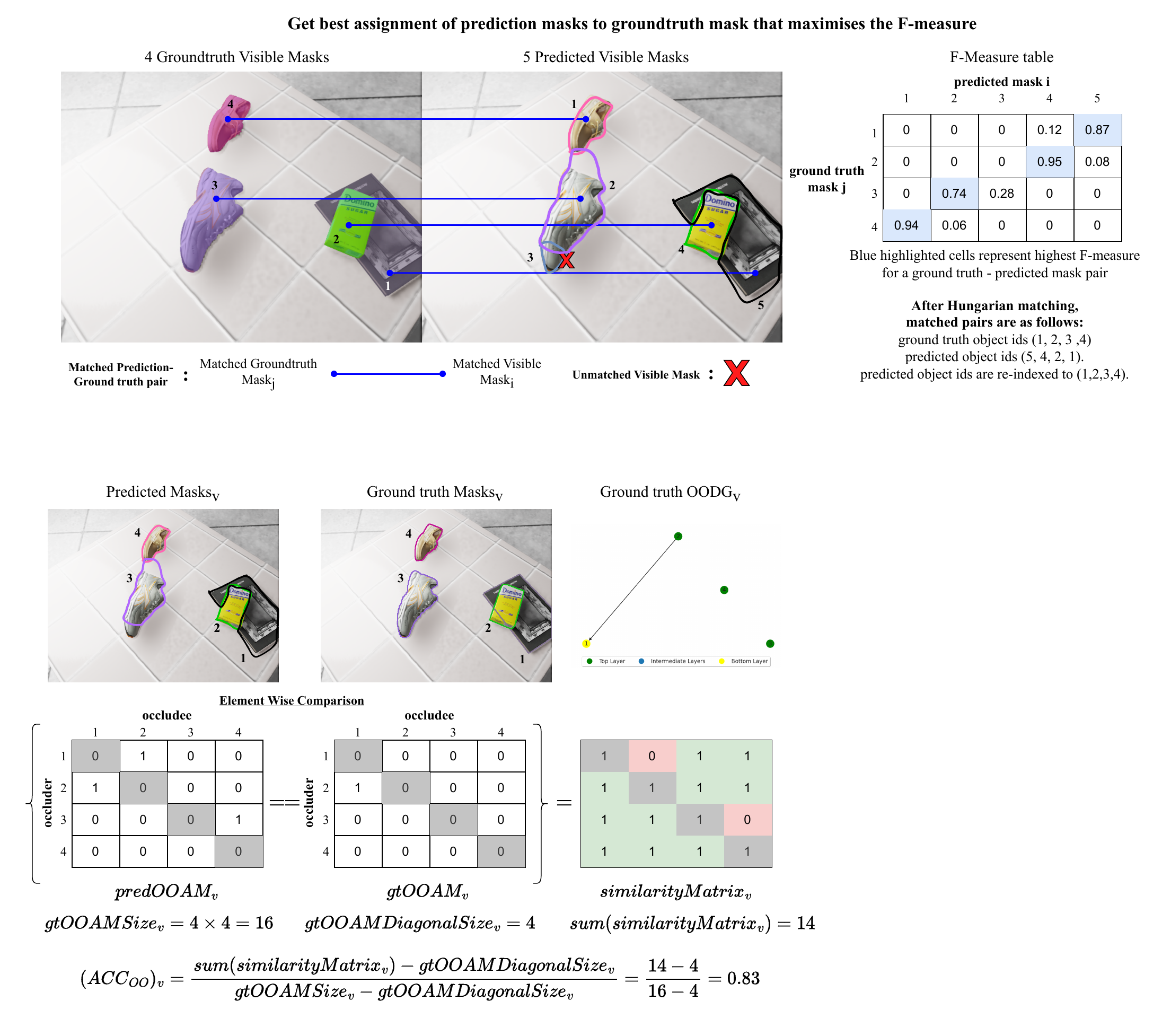}
	\caption{Hungarian Matching and calculating Occlusion Order Accuracy of image v}
	\label{hungarian_matching}	
\end{figure*}

Given an image v that depicts a typical cluttered tabletop scene, we get the ground truth-prediction assignment pairs after Hungarian matching as illustrated in Figure \ref{hungarian_matching}. The predicted masks will then be re-indexed to match the ids of the ground truth masks. Following that, the \textit{predVisible} and \textit{predOcclusion} masks that belong to the assigned pairs will be extracted. After that, the ground truth OOAM (\textit{gtOOAM}) and the predicted OOAM (\textit{predOOAM}) will be obtained using Algorithm \ref{alg:ooam}. 

Figure \ref{hungarian_matching} also illustrates the calculation of occlusion order accuracy in an image v. The similarity matrix (denoted as \textit{similarityMatrix} in Figure \ref{hungarian_matching}) is obtained by conducting an element-wise equality comparison between the \textit{gtOOAM} and \textit{predOOAM}. After that, $ACC_{oo}$ can be calculated using Equation \ref{eqn:Occlusion Order Accuracy}.


In Equation \ref{eqn:Occlusion Order Accuracy}, the $ACC_{oo}$ represents the ratio of the number of correct predicted occlusion nodes over the number of ground truth occlusion nodes. Let \textit{\#correctPredictedOcclusionNodes} denote the number of correct occluder and occludee predictions for all objects in a viewpoint (represented by green highlighted cells in \textit{similarityMatrix} in Figure \ref{hungarian_matching}).  

A summation of all the elements in the similarity matrix is carried out to obtain \textit{\#correctPredictedOcclusionNodes}. Let \textit{\#groundtruthOcclusionNodes} denote the number of ground truth occluder and occlude nodes in a viewpoint. To obtain \textit{\#groundtruthOcclusionNodes}, we count the number of elements (\textit{gtOOAMSize}) in the ground truth OOAM. As an object cannot occlude itself, the diagonal of any OOAM is always 0, and the diagonal of any similarity matrix is always 1 (depicted as grey highlighted cells in Figure \ref{hungarian_matching}). Thus, we subtract the number of elements along the diagonal of the gtOOAM (denoted by \textit
{gtOOAMDiagonalSize}) from the calculation of \textit{\#correctPredictedOcclusionNodes} and \textit{\#groundtruthOcclusionNodes}.

Correct occlusion order predictions occur when the predicted occlusion relationship for each object matches the ground truth. Incorrect occlusion order predictions can result from erroneous predictions or missing visible mask predictions of object instances. When there are missing predictions, setting the corresponding row and column of the missing object instance in the similarity matrix to 0 penalizes the model for the missing object predictions. The smaller element-wise sum of the similarity matrix leads to a smaller $ACC_{oo}$. This demonstrates the appropriate assignment of penalties by $ACC_{oo}$ to different error types for measuring object occlusion ordering in a scene.

\section{Occlusion Order Directed Acyclic Graph (OODAG)}
\label{sec5}
After obtaining the Occlusion Order Adjacency Matrix (OOAM), we can generate the occlusion order directed graph from it. For each non-zero entry ($i$, $j$) in the OOAM, we draw a directed edge from node $i$ to node $j$. If the entry is zero, we do not draw an edge. A non-zero entry at ($i$,$j$) represents that object $i$ is occluding object $j$.

\begin{figure*}[t]
	\centering
        \includegraphics[scale=0.2]{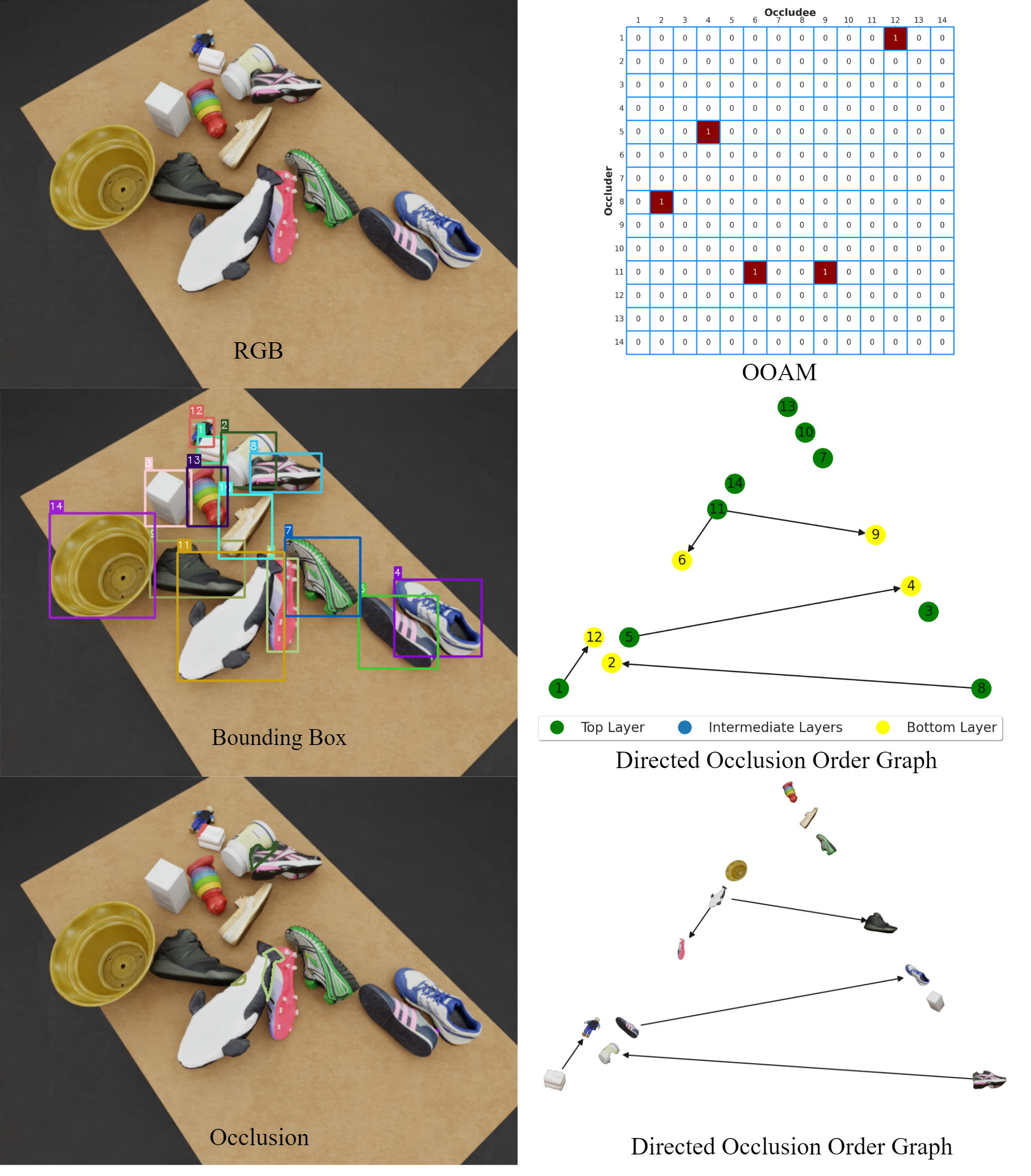}
	\caption{A visualisation of annotations for a cluttered tabletop image generated by SynTable}
	\label{oodag}	
\end{figure*}

For example, the OOAM generated in Figure \ref{oodag} shows that $(i,j)=(1,12)$ where $i$ and $j$ are the object indices (the bounding box labels) in the image. This means that object 1 occludes object 12, and a directed edge will point from object 1 to 12. From the generated Directed Occlusion Graph, we can also check if the graph is cyclic or acyclic using graph cyclic detection methods such as Depth First Search (DFS) and Breadth First Search (BFS). Only if the graph has no directed cycles (Directed Acyclic Occlusion Graph) can topological sorting be implemented.

In the generated Occlusion Order graph, we further classify objects in three different order layers - Top, Intermediate, and Bottom. Objects at the top layer represent objects that are not occluded by any other object. Objects in the intermediate layers mean that they are occluded but they also occlude other objects. For objects in the bottom layer, they are occluded but they do not occlude other objects.

\section{Qualitative Inference Results of UOAIS-Net on the OSD-Amodal Dataset}
\label{sec6}
\begin{figure*}[!ht]
	\centering
	\includegraphics[width=\linewidth]{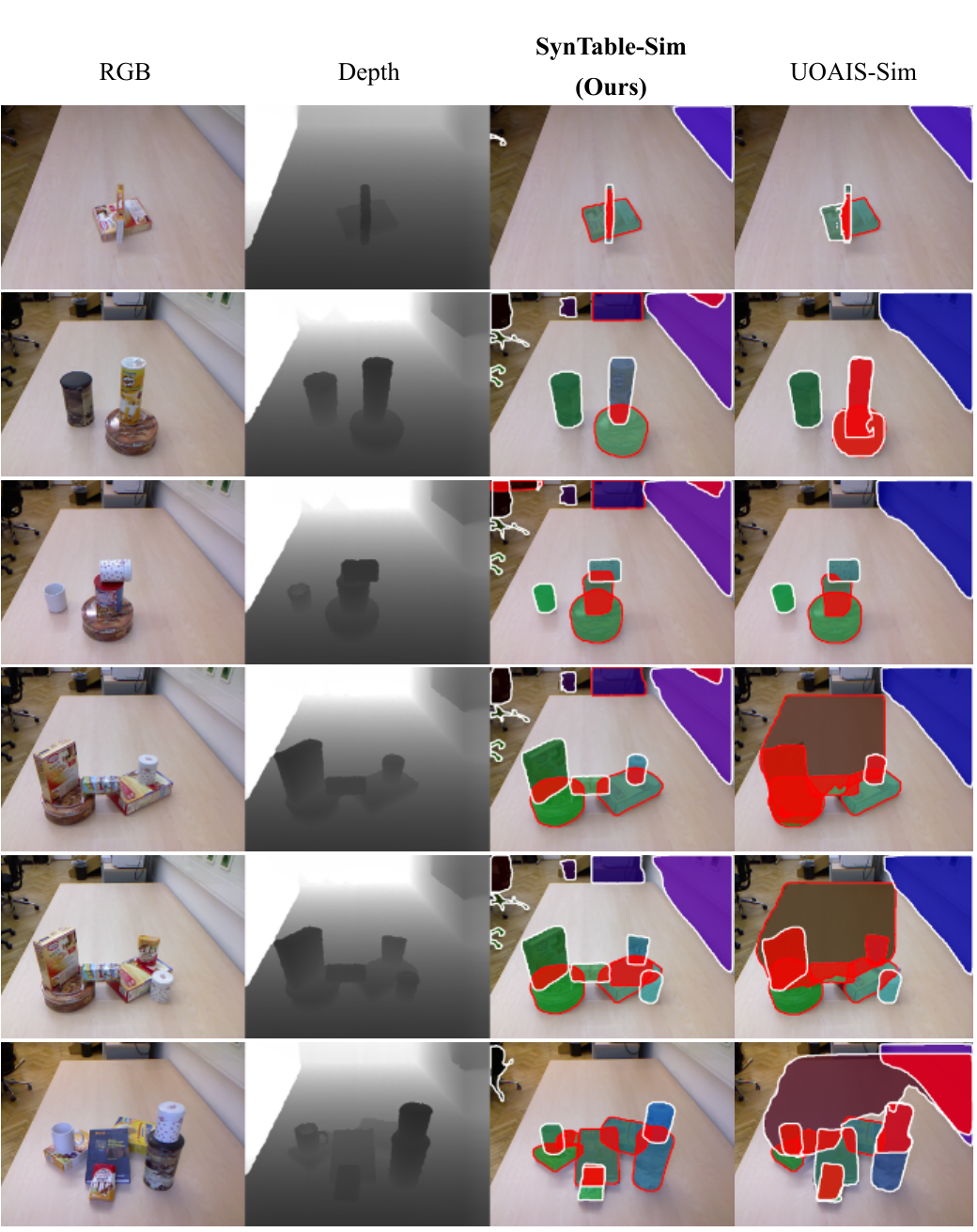}
	\caption{Comparison of the inference results on the OSD-Amodal dataset. \textbf{SynTable-Sim (Ours):} the performance of UOAIS-Net on the OSD-Amodal dataset after training on the SynTable-Sim dataset. \textbf{UOAIS-Sim}: the performance of UOAIS-Net on the OSD-Amodal dataset after training on the UOAIS-Sim tabletop dataset.}
	\label{compare}	
\end{figure*}

After training the UOAIS-Net model \cite{back2022unseen} on both SynTable-Sim and UOAIS-Sim (tabletop) datasets \cite{back2022unseen}, we present some of our qualitative results in Figure \ref{compare}. As discussed in the main text of our paper, the UOAIS-Net trained on the SynTable-Sim dataset exhibits superior performance in contrast to the UOAIS-Net trained on the UOAIS-Sim tabletop dataset. This observation is further supported by the inference results presented in Figure \ref{compare}. Furthermore, as the scene becomes more and more cluttered, the UOAIS-Net model trained on the SynTable-Sim dataset evidently outperforms that of the UOAIS-Net trained on the UOAIS-Sim tabletop dataset. 

\end{document}